%% file: arxiv.tex
\documentclass[10pt,journal,compsoc]{IEEEtran}

\usepackage{amsmath,amsfonts}
\usepackage{algorithmic}
\usepackage{algorithm}
\usepackage{array}
\usepackage[caption=false,font=footnotesize,labelfont=rm,textfont=rm]{subfig}
\usepackage{textcomp}
\usepackage{stfloats}
\usepackage{url}
\usepackage{verbatim}
\usepackage{graphicx}
\usepackage{xspace}
\usepackage{amsthm}
\usepackage{booktabs}
\usepackage{caption}
\usepackage{makecell}
\usepackage{blindtext}
\usepackage{bm}
\usepackage{bbm} 
\usepackage{comment}
\usepackage{multirow}
\usepackage{enumitem}
\usepackage{setspace}
\usepackage[safe]{tipa} 
\usepackage{color,xcolor,colortbl}
\usepackage[misc,geometry]{ifsym} 
\usepackage{pifont}
\usepackage{wrapfig}
\usepackage{ulem}
\usepackage{cancel}
\usepackage{soul}
\usepackage{ragged2e}

\usepackage{dsfont}

\usepackage{listings}
\usepackage{xcolor}
\usepackage{pythonhighlight}

\lstdefinestyle{pythonstyle}{
    language=Python,
    backgroundcolor=\color{white},   
    basicstyle=\footnotesize\ttfamily, 
    breaklines=true,                  
    numberstyle=\tiny\color{gray},
    commentstyle=\color{green},
    keywordstyle=\color{blue},
    stringstyle=\color{red},
    frame=single
}

\usepackage[hidelinks]{hyperref}

\theoremstyle{plain}

\theoremstyle{definition}

\theoremstyle{remark}

\input{math_commands.tex}

\usepackage[capitalize,noabbrev]{cleveref}
\Crefname{section}{Section}{Sections}
\crefname{section}{Sec.}{Sec.}
\Crefname{table}{Table}{Tables}
\crefname{table}{Tab.}{Tab.}
\Crefname{figure}{Figure}{Figures}
\crefname{figure}{Fig.}{Fig.}
\Crefname{equation}{Equation}{Equations}
\crefname{equation}{Eq.}{Eq.}

\definecolor{mplgreen}{rgb}{0.17254901960784313, 0.6274509803921569, 0.17254901960784313}
\definecolor{myblue}{rgb}{0.2, 0.2, 0.9}
\definecolor{myred}{rgb}{1.0, 0.1, 0.1}

\urldef\rrcurl\url{https://pytorch.org/vision/stable/generated/torchvision.transforms.RandomResizedCrop.html}
\urldef\imageneturl\url{https://storage.googleapis.com/bit_models/imagenet21k_wordnet_lemmas.txt}
\urldef\cifarurl\url{https://www.cs.toronto.edu/%7Ekriz/cifar.html}
\urldef\adaptformerurl\url{https://github.com/ShoufaChen/AdaptFormer/blob/main/models/adapter.py}

\newcolumntype{L}[1]{>{\raggedright\arraybackslash}p{#1}}
\newcolumntype{R}[1]{>{\raggedleft\arraybackslash}p{#1}}
\newcolumntype{C}[1]{>{\centering\arraybackslash}p{#1}}

\captionsetup[figure]{aboveskip=0.05in, belowskip=0.05in}
\captionsetup[table]{aboveskip=0.05in, belowskip=0.05in}

\allowdisplaybreaks

\ifCLASSOPTIONcompsoc
  \usepackage[sort,compress]{cite}
\else
  \usepackage[sort,compress]{cite}  

\fi
\ifCLASSINFOpdf
\else
\fi

\hyphenation{op-tical net-works semi-conduc-tor IEEE-Xplore}

\begin{document}
%
\title{Sparsity Hurts: Simple Linear Adapter Can Boost Generalized Category Discovery}

\author{Bo~Ye,
        Kai~Gan,
        Tong~Wei,
        and~Min-Ling~Zhang,~\IEEEmembership{Senior Member,~IEEE}
\IEEEcompsocitemizethanks{\IEEEcompsocthanksitem Bo~Ye, Kai~Gan, Tong~Wei, and Min-Ling~Zhang are with the School of Computer
Science and Engineering, Southeast University, Nanjing 210096, China,
and the Key Laboratory of Computer Network and Information Integration (Southeast University), Ministry of Education, China. E-mail:
\{yeb, gank, weit, zhangml\}@seu.edu.cn.}

}
%
%

\markboth{Journal of \LaTeX\ Class Files,~Vol.~XX, No.~XX, AUGUST~XX}%
{Shell \MakeLowercase{\textit{et al.}}: Bare Demo of IEEEtran.cls for Computer Society Journals}

%


\IEEEtitleabstractindextext{%
\begin{abstract}
\justifying\let\raggedright\justifying
Generalized Category Discovery (GCD) seeks to identify novel categories from unlabeled data while retaining the classification ability of seen categories. Prior GCD methods commonly leverage transferable representations from pre-trained models, adapting to downstream datasets via partial fine-tuning (updating only the final ViT block) and visual prompt tuning (appending learnable vectors to inputs). However, conventional partial fine-tuning offers limited flexibility, as it fails to adapt the entire model; meanwhile, visual prompt tuning is prone to overfitting, due to its sensitivity to initialization and inherently constrained capacity.
To address these limitations, we propose {LAGCD}, a simple yet effective GCD approach that embeds a residual linear adapter into each ViT block. From the perspective of feature sparsity, we systematically show that non-linearity in conventional adapters impairs performance, whereas our linear adapter enhances it by enabling more flexible model capacity. We further introduce an auxiliary distribution alignment loss to mitigate the negative impact of biased predictions between seen and novel categories. Extensive experiments on both generic and fine-grained datasets confirm that LAGCD consistently improves performance over many sophisticated baselines. The source code is available at \url{https://github.com/yebo0216best/LAGCD}.
\end{abstract}


\begin{IEEEkeywords}
Generalized category discovery, semi-supervised learning, novel class discovery, parameter-efficient fine-tuning.
\end{IEEEkeywords}}

\let\oldtwocolumn\twocolumn
\renewcommand\twocolumn[1][]{%
\oldtwocolumn[{#1}{
    \vspace{-0.3in}
    \centering
    \includegraphics[width=0.47\linewidth]{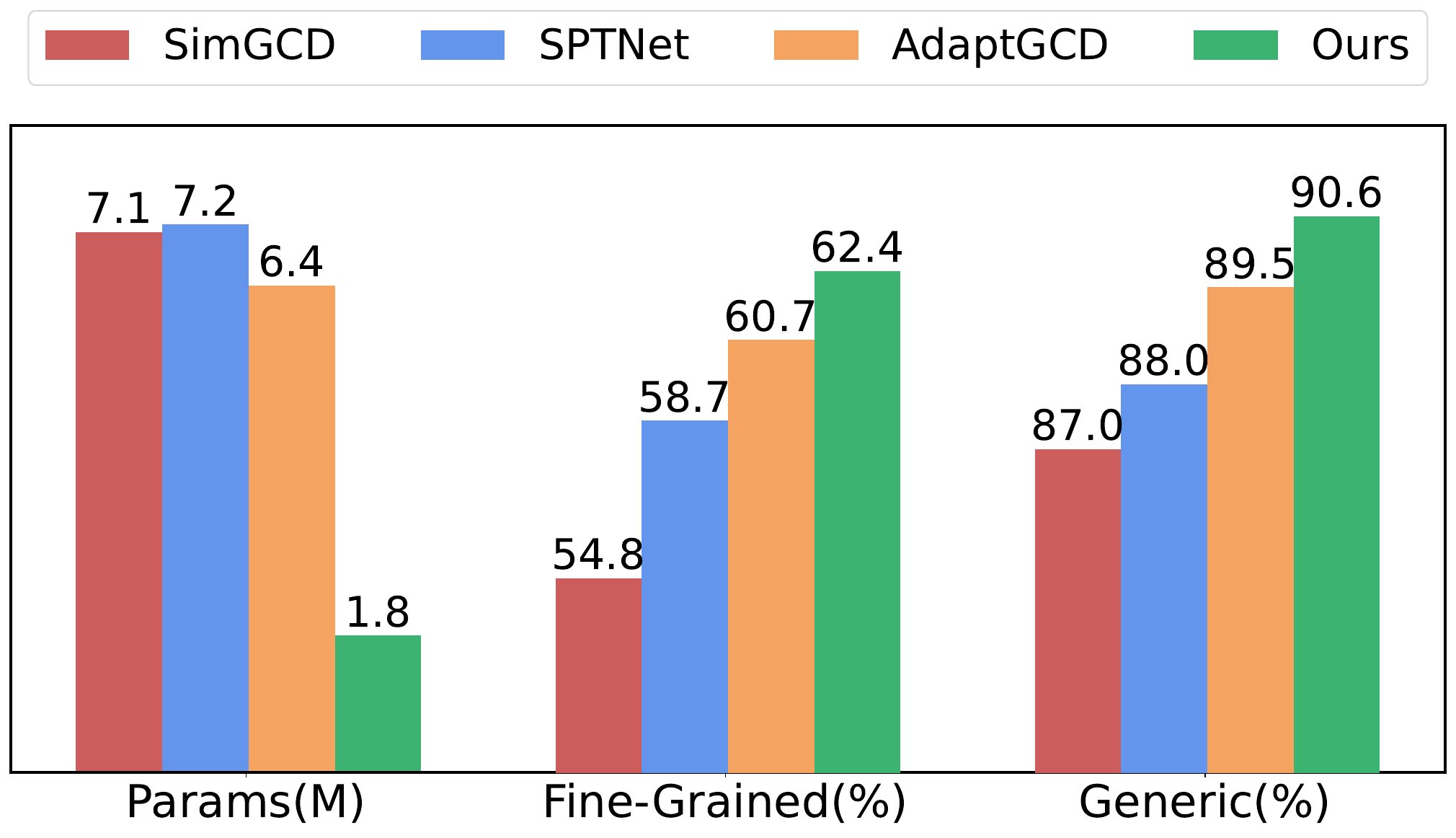}
    \hfill
    \includegraphics[width=0.525\linewidth]{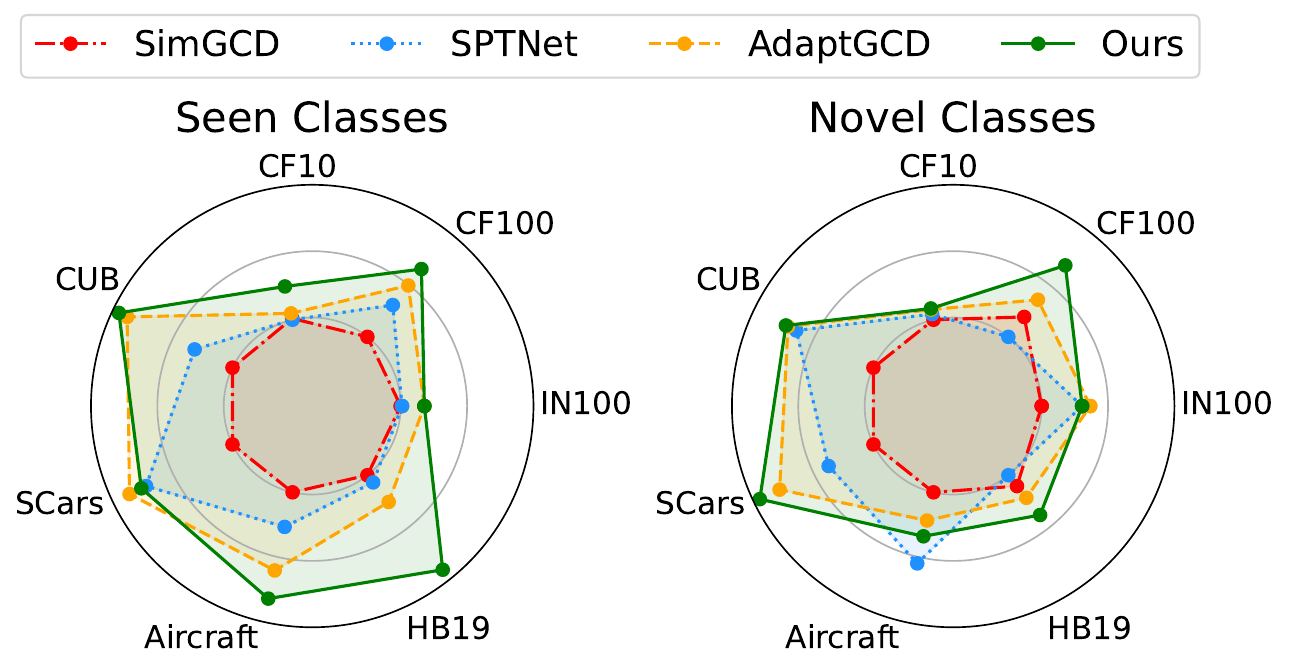}
    \captionof{figure}{\textbf{Performance comparisons with SimGCD~\cite{wen2023parametric}, SPTNet~\cite{wang2024sptnet}, and AdaptGCD~\cite{qu2024adaptgcd}.} (a) We compare the number of tunable parameters of the ViT backbone, and accuracy averaged over four fine-grained and three generic datasets. (b) Seen and novel class accuracy of our method and existing methods on seven datasets.}
    \vspace{0.3in}
    \label{fig:zzt_comparison}
}]
}

\maketitle

\IEEEdisplaynontitleabstractindextext

\IEEEpeerreviewmaketitle

\section{Introduction}\label{sec:introduction}

\IEEEPARstart{R}{ecently},  semi-supervised learning (SSL)~\cite{lee2013pseudo,sajjadi2016regularization,laine2016temporal,sohn2020fixmatch,wei2023towards,ganerasing,gansemi} has emerged as a promising paradigm for leveraging large-scale unlabeled data. However, it struggles when unlabeled data contain novel categories which are not seen in labeled data, violating its core assumption of complete category coverage and hindering its ability to mining useful information from novel categories~\cite{oliver2018realistic, cao2022openworld, vaze2022generalized}. To tackle this, Generalized Category Discovery (GCD) has emerged as a widely studied topic in recent research~\cite{cao2022openworld, vaze2022generalized,wen2023parametric,zhang2023promptcal,sun2023graph,wang2024sptnet,qu2024adaptgcd}. GCD tackles the dual task of identifying novel categories from unlabeled data while maintaining classification capability on seen categories. The limited supervision from seen categories, coupled with the absence of accurate information about novel categories, makes it difficult for models to generalize effectively.

Existing GCD methods~\cite{vaze2022generalized, wen2023parametric, vaze2024no} utilize DINO-pretrained Vision Transformer (ViT)~\cite{dosovitskiy2020image, caron2021emerging} as backbone to extract transferable semantic representations, while addressing overfitting by adopting an effective partial fine-tuning strategy, wherein only the final ViT block is updated. To improve the flexibility of the backbone, several methods~\cite{zhang2023promptcal, wang2024sptnet} have built upon Visual Prompt Tuning (VPT)~\cite{jia2022visual} as an enhancement over conventional partial fine-tuning. However, these methods reveal that direct adaptation of the backbone via visual prompts can induce training instability and degrade performance, highlighting the need for specialized mitigation strategies. For example, PromptCAL~\cite{zhang2023promptcal} introduces a warm-up phase with discriminative prompt regularization to adapt the class token and visual prompts for downstream datasets, while SPTNet~\cite{wang2024sptnet} employs a two-stage training paradigm that alternates between optimizing model parameters and visual prompts. Unlike fine-tuning paradigms discussed above, adapter-based tuning~\cite{houlsby2019parameter, hu2021lora, chen2022adaptformer} still remains underexplored in GCD~\cite{qu2024adaptgcd}. The paradigm utilizes a two-layer \texttt{down}$\rightarrow$\texttt{up} bottleneck module with zero-initialized up-projection layers, ensuring new parameters initially approximate the behavior of the backbone. Additionally, AdaptFormer~\cite{chen2022adaptformer} demonstrates that adapter-based tuning achieves greater stability when scaling up trainable parameters by expanding bottleneck dimensions, compared to VPT, which relies on increasing the number of visual prompts. AdaptGCD~\cite{qu2024adaptgcd} pioneers the use of adapters for GCD. To address bias between seen and novel categories, it extends SimGCD~\cite{wen2023parametric} with a multi-expert adapter structure, where a routing layer assigns samples to experts specialized in seen or novel categories. However, this design introduces high computational overhead, requiring multiple adapters per ViT block and additional loss terms for expert balancing, potentially limiting applicability and generalization.

In this paper, we propose a simple yet effective adapter-based tuning method for GCD. Instead of introducing complex architectural changes or focusing on tuning hyperparameters, we investigate the crucial yet often neglected role of ReLU activation~\cite{nair2010rectified} within adapters~\cite{chen2022adaptformer}. Our analysis reveals that ReLU-induced sparsity could discard important information and further impair effective adaptation. Through extensive validation experiments, we find that a simple linear transformation by removing the activation achieves robust generalization, suggesting that the non-linearity and sparsity introduced by ReLU are not only unnecessary but may even hinder performance. We further observe that varying adapter hyperparameters yields stable performance for seen categories, driven by rapidly learned discriminative features, but leads to notable fluctuations on novel categories. In addition, directly applying adapter without ReLU activation still incurs prediction bias due to the inherent learning discrepancies between seen and novel categories~\cite{cao2022openworld,wen2023parametric}. Moreover, the balanced sampling commonly used to balance supervised and unsupervised contrastive learning tends to oversample labeled data while undersampling unlabeled data, potentially exacerbating the bias. To alleviate this, we optimize an auxiliary distribution alignment loss without compromising contrastive representation learning. As shown in Figure~\ref{fig:zzt_comparison}, our method outperforms the baseline SimGCD~\cite{wen2023parametric} as well as the prompt-based method SPTNet~\cite{wang2024sptnet} in terms of accuracy, while requiring fewer tunable parameters, with both methods relying on partial fine-tuning and SPTNet additionally incorporating visual prompts. Compared with the adapter-based method AdaptGCD~\cite{qu2024adaptgcd}, our method achieves superior performance with fewer trainable parameters.

\textbf{Our contributions are as follows:} (1) We systematically analyze the negative effects of feature sparsity caused by nonlinear activation function in adapters, which is overlooked in prior literature;  {(2)} We propose LAGCD, a simple but effective GCD method by removing the activation function within adapters, coupled with an auxiliary distribution alignment to ensure balanced predictions; {(3)} Our proposed method achieves state-of-the-art results across both generic and fine-grained benchmarks, establishing a strong baseline for future research in this field.

\section{Related Work}
\smallskip
\noindent\textbf{Generalized Category Discovery.} GCD aims to detect novel categories in unlabeled data while retaining performance on seen categories~\cite{vaze2022generalized, cao2022openworld}. Current GCD methods can be divided into non-parametric~\cite{vaze2022generalized, fei2022xcon, zhang2023promptcal,sun2022opencon,sun2023graph, choi2024contrastive} and parametric classifiers~\cite{wen2023parametric, wang2024sptnet, cao2024solving, qu2024adaptgcd, lin2024flipped, wang2024hilo, ye2024bridging}. Non-parametric classifier-based methods often combine semi-supervised contrastive learning for representation learning with semi-supervised $k$-means for category assignment. SimGCD~\cite{wen2023parametric} tackles non-parametric classifier overhead with a parametric prototype classifier and entropy maximization, improving performance and efficiency. Traditional GCD methods lean on partial fine-tuning. PromptCAL~\cite{zhang2023promptcal} and SPTNet~\cite{wang2024sptnet} propose two-stage frameworks blending partial fine-tuning with visual prompts for better adaptation. AdaptGCD~\cite{qu2024adaptgcd} is the first to introduce adapters~\cite{chen2022adaptformer} into GCD, proposing a multi-expert adapter structure. Diffierent from them, we explore how adapters can achieve the better adaptation by analyzing their activations.

\smallskip
\noindent\textbf{Adapter-based Tuning.} Fully fine-tuning large pretrained models can lead to training instability and limited transferability. Recently, adapter-based tuning methods~\cite{houlsby2019parameter, hu2021lora, chen2022adaptformer} have been proposed to address these challenges. These methods insert lightweight adapters, which are two-layer bottleneck-structured MLPs with residual connections, into pretrained models. Adapter~\cite{houlsby2019parameter}, LoRA \cite{hu2021lora}, AdaptFormer~\cite{chen2022adaptformer} are three representative adapters across various downstream tasks. Adapter~\cite{houlsby2019parameter} sequentially inserts two adapters into each transformer block: one after the Multi-head Self-Attention (MSA) and one after the MLP, adding few tunable parameters but matching the performance of full fine-tuning. LoRA~\cite{hu2021lora} incorporates trainable low-rank decomposition matrices in parallel with the original weights in the MSA, achieving competitive performance without introducing extra inference latency. AdaptFormer~\cite{chen2022adaptformer} modifies the original sequential Adapter design by introducing a scalable Adapter in parallel with the MLP, demonstrating notable effectiveness in vision downstream tasks. Inspired by the effectiveness of AdaptFormer in downstream tasks~\cite{shi2024longtail, qu2024adaptgcd}, we adopt it as our base adapter and conduct a detailed investigation into the role of activation functions within its architecture.

\section{Preliminary}
\subsection{Problem Formulation}
Let $\mathcal{D}$ denote the whole dataset, which consists of two subsets: a labeled set $\mathcal{D}_l=\{ (x_i, y_i)\}_{i=1}^{N_l} \subset \mathcal{X}_l \times \mathcal{Y}_l$ and an unlabeled set $\mathcal{D}_u=\{x_j\}_{j=N_l+1}^{N_l+N_u} \subset \mathcal{X}_u$, where $x$ represents an image sample and $y$ represents a label. The label space for the labeled and unlabeled samples is given by $\mathcal{Y}_l = \mathcal{C}_s$ and $\mathcal{Y}_u = \mathcal{C} = \mathcal{C}_s \cup \mathcal{C}_n$, where $\mathcal{C}$ is the set of all categories, and $\mathcal{C}_s$ and $\mathcal{C}_n$ represent the label sets for the \textit{Seen}, and \textit{Novel} categories, respectively. The objective of the GCD task is to categorize all the unlabeled images in $\mathcal{D}_u$, specifically by classifying samples from the seen category set $\mathcal{C}_s$ and clustering samples from the novel category set $\mathcal{C}_n$.

\subsection{Baseline: SimGCD}
SimGCD~\cite{wen2023parametric} is a representative baseline, containing two key components. One focuses on representation learning, while the other targets parametric prototype classification. 

\smallskip
\noindent\textbf{Representation Learning.} The representation learning objective of SimGCD is based on GCD~\cite{vaze2022generalized}, which incorporates supervised contrastive learning~\cite{khosla2020supervised} for labeled samples and self-supervised contrastive learning~\cite{chen2020simple} for all samples. 
Formally, given two augmented views, $\boldsymbol{x}_i$ and $\boldsymbol{x}_i^{\prime}$, of the same image in a mini-batch $B$, the self-supervised contrastive loss is defined as follows:
\begin{equation}
\mathcal{L}^u_\text{rep}=\frac{1}{|B|} \sum_{i \in B}-\log \frac{\exp \left(\boldsymbol{z}_i^\top \boldsymbol{z}_i^{\prime} / \tau_u\right)}{\sum_n^{n \neq i} \exp \left(\boldsymbol{z}_i^\top \boldsymbol{z}_n^{\prime} / \tau_u\right)}
\end{equation}
where the feature $\boldsymbol{z}_i = g\left(f\left(\boldsymbol{x}_i\right)\right)$ is $\ell_2$-normalized, with $f$ and $g$ denoting the backbone and the projection head, respectively, and $\tau_u$ being the temperature value. The supervised contrastive loss $\mathcal{L}^s_\text{rep}$ is similar, with the key difference being that positive samples are matched according to their labels. The overall representation learning loss is balanced with $\lambda_{sup}$: $\mathcal{L}_\text{rep} = (1 - \lambda_{sup}) \mathcal{L}^u_\text{rep} + \lambda_{sup} \mathcal{L}^s_\text{rep}$.

\smallskip
\noindent\textbf{Parametric Classification.} The parametric classification of SimGCD follows self-distillation~\cite{caron2021emerging, assran2022masked}. With \(K = |\mathcal{Y}_l \cup \mathcal{Y}_u|\) categories, a set of prototypes \(\mathcal{C} = \{\boldsymbol{c}_1, \dots, \boldsymbol{c}_K\}\) is randomly initialized. SimGCD calculate the probability for each augmented view \(\boldsymbol{x}_i\) by applying softmax on the cosine similarity between the hidden feature \(\boldsymbol{h}_i = f(\boldsymbol{x}_i)\) and prototypes:
\begin{equation}
\label{eq:output_prob}
\boldsymbol{p}_{i}^{(k)}=\frac{\exp \left(\frac{1}{\tau_s}(\boldsymbol{h}_i/||\boldsymbol{h}_i||_2)^\top (\boldsymbol{c}_k / ||\boldsymbol{c}_k||_2)\right)}{\sum_{k^\prime} \exp \left(\frac{1}{\tau_s}(\boldsymbol{h}_i / ||\boldsymbol{h}_i||_2)^\top (\boldsymbol{c}_{k^\prime} / ||\boldsymbol{c}_{k^\prime}||_2)\right)}
\end{equation}
where $\tau_s$ is a hyperparameter that adjusts the `student' probability, and $k$ denotes the $k$-th class. The classification objective is the cross-entropy $\ell(\boldsymbol{y}, \boldsymbol{p}) = -\sum_{k} {\boldsymbol{y}^{(k)}}\log{\boldsymbol{p}}^{(k)}$. For labeled data, $\mathcal{L}^s_\text{cls}$ is computed directly using their one-hot ground-truth labels, while for unlabeled samples, we derive soft pseudo-labels $\boldsymbol{q}_i^\prime$ from another view $\boldsymbol{x}_i^{\prime}$ using a sharper temperature $\tau_t$ than $\tau_s$, as defined in Eq.~\ref{eq:output_prob}. The supervised and unsupervised losses are:
\begin{equation} \label{eq:selfdistill1}
     \mathcal{L}^s_\text{cls} = \frac{1}{|B^l|}\sum_{i \in B^l}\ell(\boldsymbol{y}_i, \boldsymbol{p}_i)
\end{equation}
\begin{equation} \label{eq:selfdistill2}
    \mathcal{L}^u_\text{cls} = \frac{1}{|B|}\sum_{i \in B}\ell(\boldsymbol{q}_i^\prime, \boldsymbol{p}_i) - \lambda_{ent} H(\overline{\boldsymbol{p}})
\end{equation}
where $\boldsymbol{y}_i$ denotes the one-hot label of $\boldsymbol{x}_i$. $H(\overline{\boldsymbol{p}})$ is the maximum entropy regularization~\cite{Arazo2020ijcnn, assran2022masked}, and $\lambda_{ent}$ is the weight hyperparameter. The detailed form of $H(\overline{\boldsymbol{p}})$ is:
\begin{equation}
    H(\overline{\boldsymbol{p}}) = -\sum_k\overline{\boldsymbol{p}}^{(k)}\log\overline{\boldsymbol{p}}^{(k)}
\end{equation}
where $\overline{\boldsymbol{p}} = \frac{1}{2|B|}\sum_{i \in B}\left( \boldsymbol{p}_i+\boldsymbol{p}_i^\prime \right)$ denotes the mean predicted probability of a batch. The total classification objective is $\mathcal{L}_\text{cls} = (1 - \lambda_{sup}) \mathcal{L}^u_\text{cls} + \lambda_{sup} \mathcal{L}^s_\text{cls}$, where $\lambda_{sup}$ is the same as $\lambda_{sup}$ of $\mathcal{L}_\text{rep}$, and the overall objective of SimGCD is simply $\mathcal{L}_\text{rep} + \mathcal{L}_\text{cls}$.

\subsection{Base Adapter: AdaptFormer}
In this paper, we primarily investigate enhancements based on AdaptFormer~\cite{chen2022adaptformer} due to its simplicity and effectiveness. AdaptFormer introduces an additional branch within the MLP block of ViT, which is a lightweight module designed for task-specific fine-tuning. Specifically, the branch is designed as a bottleneck structure, comprising a down-projection layer with parameters $\bm{W}_{\rm down} \in \mathbb{R}^{d \times \hat{d}}$ and $\textbf{b}_{\rm down} \in \mathbb{R}^{\hat{d}}$, and an up-projection layer with parameters $\bm{W}_{\rm up} \in \mathbb{R}^{\hat{d} \times d}$ and $\textbf{b}_{\rm up} \in \mathbb{R}^{d}$, where $\hat{d}$ represents the bottleneck dimension. It also incorporates the ReLU activation function between $\bm{W}_{\rm down}$ and $\bm{W}_{\rm up}$ for non-linearity. And the branch is integrated with the original MLP via residual connection, scaled by a factor $\mathit{s}_a$. Formally, given an input feature $\boldsymbol{x'}_\ell$, the additional branch generates the adapted feature $\boldsymbol{\tilde{x}}_\ell$:
\begin{equation}
    \label{cal::bottleneck}
    \begin{aligned}
        \boldsymbol{\tilde{x}}_\ell = {\rm ReLU}({\rm LN}(\boldsymbol{x'}_\ell)\cdot\textbf{W}_{\rm down}^{\top} + \textbf{b}_{\rm down})\cdot\textbf{W}_{\rm up}^{\top} + \textbf{b}_{\rm up}
    \end{aligned}
\end{equation}
Subsequently, both the adapted feature $ \boldsymbol{\tilde{x}}_\ell$ and MLP feature are fused with the input feature $\boldsymbol{x'}_\ell$ via a residual connection:
\vspace{-1pt}
\begin{equation}
    \boldsymbol{x}_\ell =  {\rm MLP}({\rm LN}(\boldsymbol{x'}_\ell)) + \boldsymbol{x'}_\ell + \boldsymbol{\tilde{x}}_\ell\cdot\mathit{s}_a
    \label{eq:adapter}
\end{equation}
where $\mathit{s}_a$ is the scaling factor, and LN($\cdot$) denotes the layer normalization. During fine-tuning, only the parameters of the bottleneck modules are updated and the rest are frozen.

\section{Method}

\subsection{Analyzing the Sparsity of Adapter}

In GCD, models often exhibit pronounced bias fluctuations during training, as labeled data primarily drives learning of the seen categories and only indirectly supports generalization to novel ones~\cite{cao2022openworld,wen2023parametric}. This intrinsic challenge imposes substantial difficulties for effective adaptation. When tackling complex downstream tasks, optimizing models with adapters is largely influenced by the choice of the bottleneck dimension $\hat{d}$ and the scaling factor $\mathit{s}_a$. Intuitively, excessively large values may introduce training instability and overfitting, whereas overly small values can result in insufficient adaptation to downstream datasets.

In this work, we examine a crucial yet often neglected component in adapters: the intrinsic ReLU activation function, which introduces non-linearity and sparsity to enhance performance~\cite{chen2022adaptformer}. Rather than emphasizing the tuning of the bottleneck dimension $\hat{d}$ and the scaling factor $\mathit{s}_a$ in adapters, we instead examine how activation functions influence model adaptation. Based on the evaluation of ReLU and its variants in deep neural networks, Xu et al.~\cite{xu2015empirical} demonstrated that introducing a non-zero slope for negative inputs in ReLU consistently improves performance. Intuitively, ReLU-induced sparsity through zeroing negative values may discard important information during feature propagation, potentially constraining model adaptation. Motivated by this consideration, we perform a systematic analysis of ReLU's role in the adapter-based GCD setting and address a central question: \textit{Does ReLU-induced non-linearity or sparsity within adapters truly enhance their adaptability?}


\textbf{Analysis Methods: }
To quantify the sparsity of ReLU and analyze its impact more effectively, we design experiments from three perspectives: the activation gradient perspective, where Leaky-ReLU is used to adjust the gradient magnitude of negative values; the activation range perspective, where Threshold-ReLU is employed to directly control the extent of zero-valued outputs; and the activation distribution perspective, where ELU is adopted to introduce smooth negative activations and alleviate the strict sparsity.

\textbf{Leaky-ReLU}~\cite{maas2013rectifier} introduces a small fixed non-zero slope $\alpha$ (with the special case of $0\leq\alpha\leq1$ in our analysis) to provide the tunable gradient for negative inputs, which transforms ReLU's inherently sparse activation pattern into a dense pattern. Leaky-ReLU generalizes activation behaviors: as $\alpha$ approaches 0, it converges to ReLU, while as $\alpha$ approaches 1, it converges to linear transformation. Formally, Leaky-ReLU is defined as follows:

\begin{equation}
\mathrm{Leaky\mbox{-}ReLU}(x) =
\begin{cases}
x, & x \geq 0 \\
\alpha x, & x < 0
\end{cases}
\end{equation}

\textbf{Threshold-ReLU}~\cite{konda2014zero} introduces a fixed threshold $t$ to replace the zero-threshold mechanism in ReLU. 
This modification makes the sparsity of ReLU adjustable, which still maintains its sparse nature and is more intuitive compared to Leaky-ReLU. The threshold $t$ controls sparsity levels with improved interpretability: higher $t$ values force more activations to zero, creating greater sparsity. Specifically, Threshold-ReLU is defined as follows:

\begin{equation}
\mathrm{Threshold\mbox{-}ReLU}(x) =
\begin{cases}
x, & x \geq t \\
0, & x < t
\end{cases}
\end{equation}

\textbf{ELU (Exponential Linear Unit)}~\cite{clevert2015fast} extends ReLU by allowing negative outputs in an exponential form, which alleviates strict sparsity and shifts the mean activation closer to zero, thereby facilitating gradient flow during training. For large negative inputs, ELU saturates at a fixed value $-\beta (\beta>0)$. A larger $\beta$ amplifies negative responses and reduces sparsity, whereas as $\beta$ approaches 0, ELU degenerates to ReLU. Formally, ELU is defined as:

\begin{equation}
\mathrm{ELU}(x) =
\begin{cases}
x, & x \geq 0 \\
\beta (e^x - 1), & x < 0
\end{cases}
\end{equation}

\begin{figure}[!h]
\centering
\includegraphics[width=0.98\linewidth]{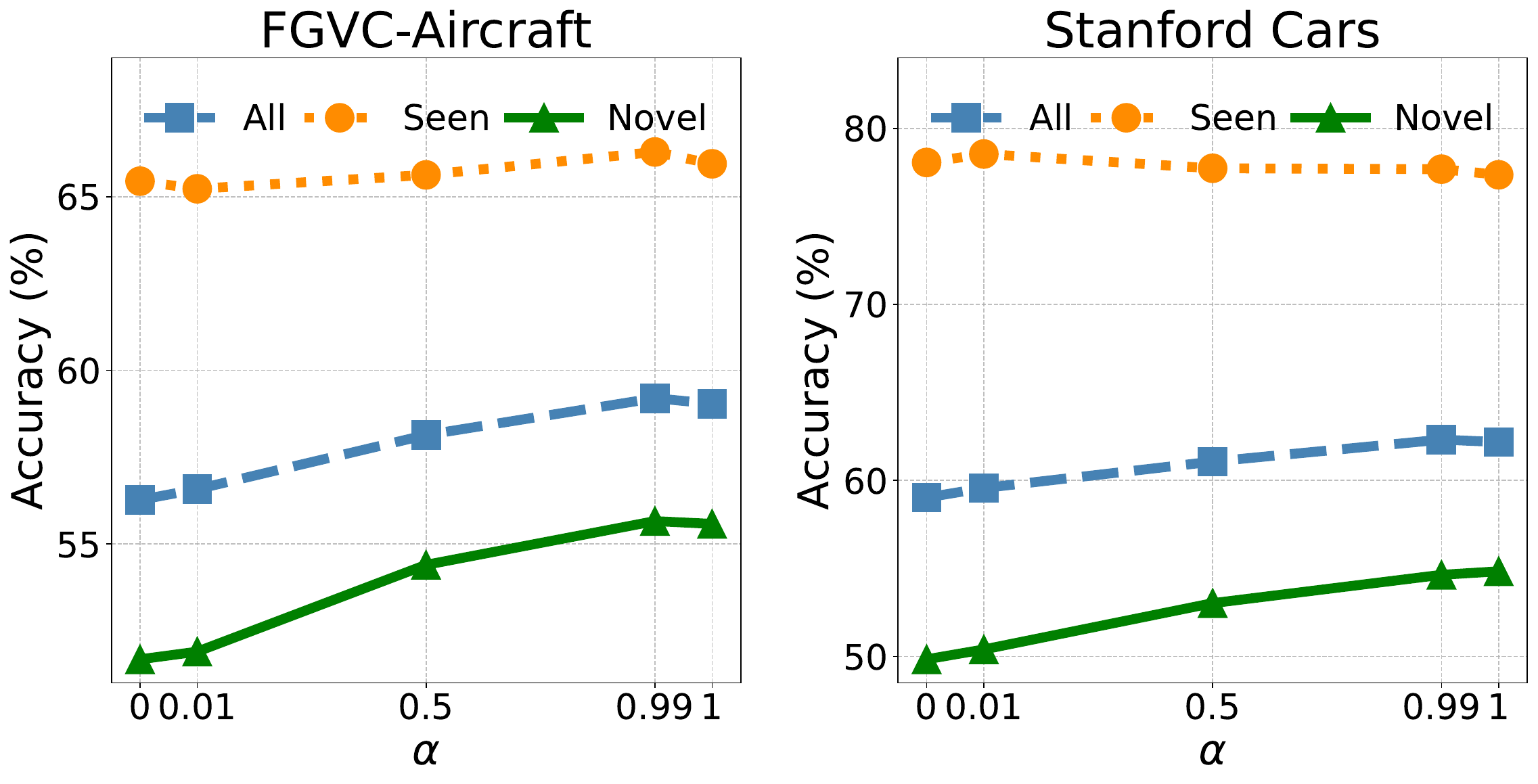}
\caption{Different slopes $\alpha$ in Leaky-ReLU.}
\label{fig:alpha}
\end{figure}

\begin{figure}[!h]
\centering
\includegraphics[width=0.98\linewidth]{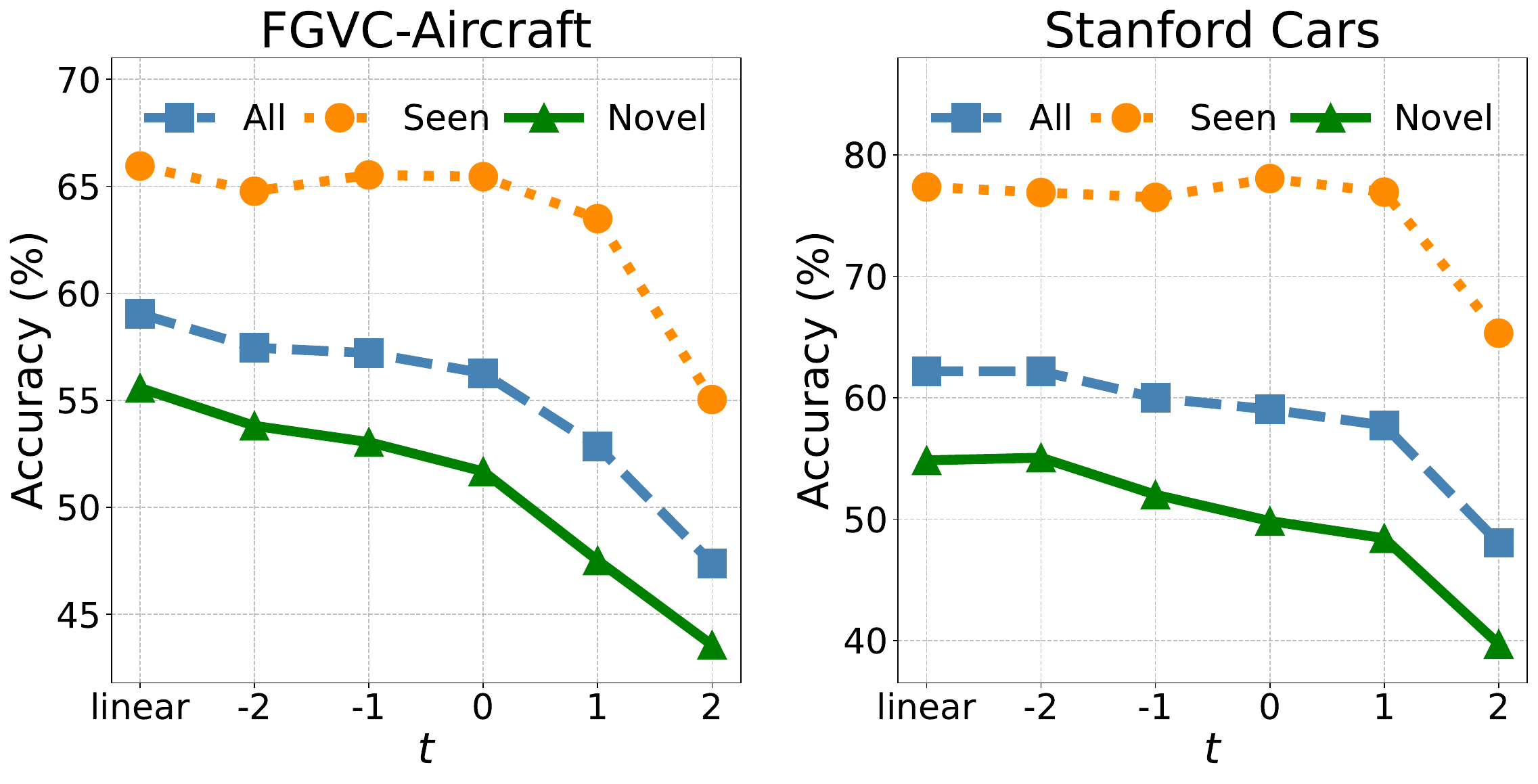}
\caption{Different thresholds $t$ in Threshold-ReLU.}
\label{fig:t}
\end{figure}

\begin{figure}[!h]
\centering
\includegraphics[width=0.98\linewidth]{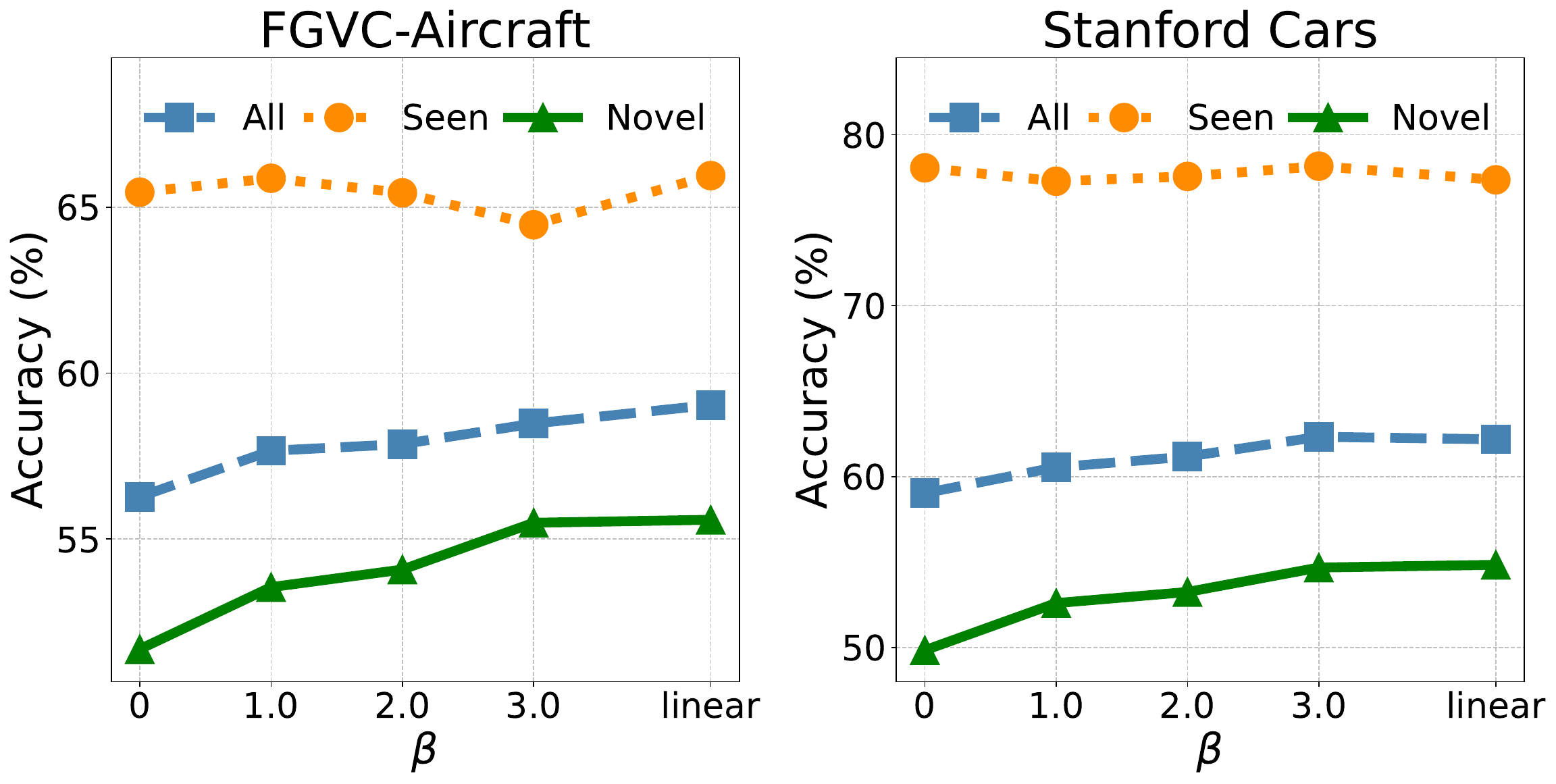}
\caption{Different slopes $\beta$ in ELU.}
\label{fig:beta}
\end{figure}

For experimental validation, we employ SimGCD~\cite{wen2023parametric} with adapters on two challenging fine-grained datasets FGVC-Aircraft~\cite{maji2013fine} and Stanford Cars~\cite{krause20133d}, where ReLU is replaced with Leaky-ReLU, Threshold-ReLU, or ELU. We explore different activation settings by varying the slope $\alpha \in \{0, 0.01, 0.5, 0.99, 1\}$, the threshold $t \in \{-2, -1, 0, 1, 2\}$, and the slope $\beta \in \{0, 1, 2, 3\}$, while keeping all other hyperparameters consistent. In addition, to examine whether different adapters induce overfitting or underfitting, we compute feature similarity for each sample before and after fine-tuning the pre-trained model.

\begin{figure}[!h]
\centering
\includegraphics[width=0.98\linewidth]{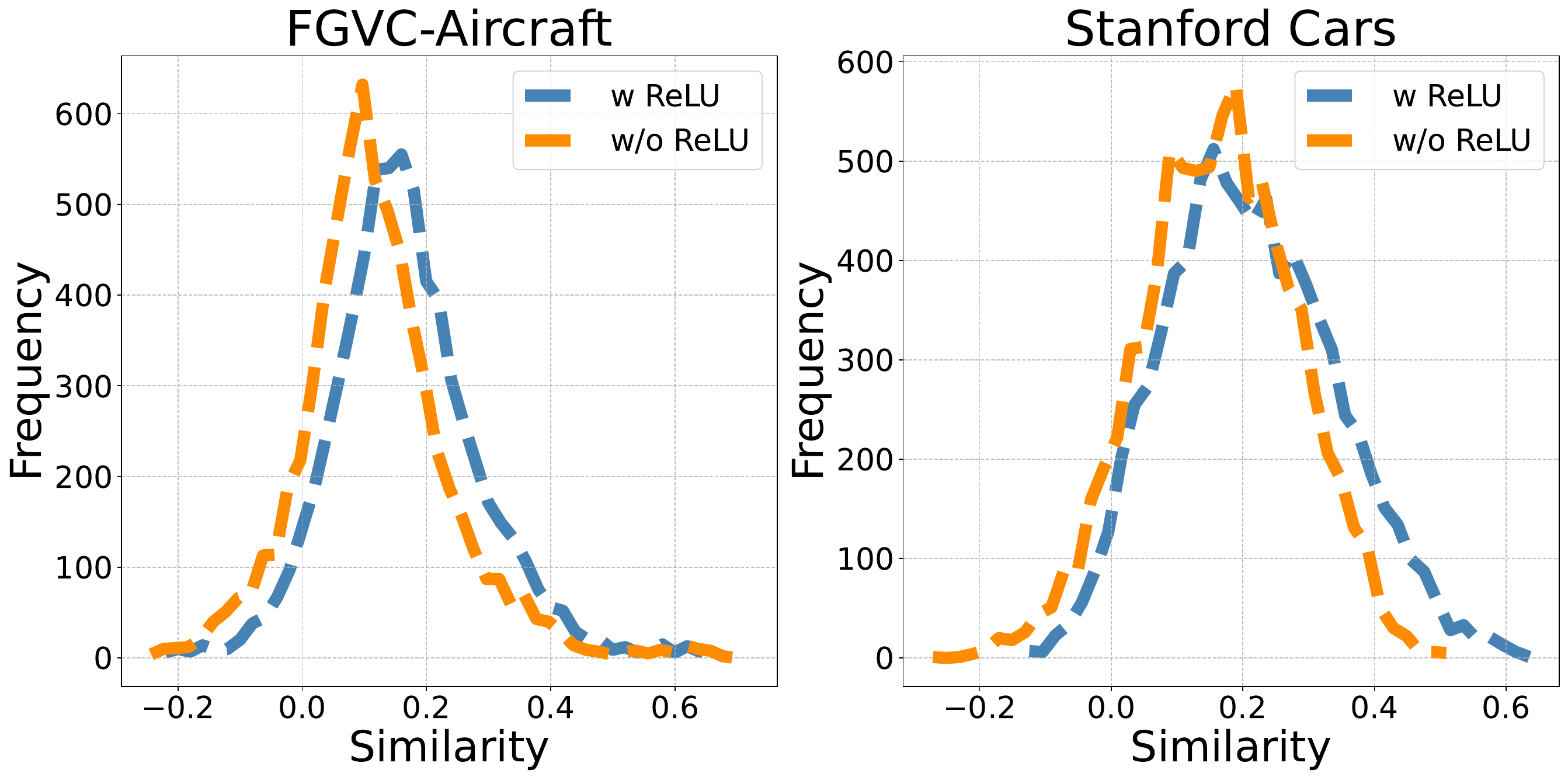}
\caption{\textbf{Similarity between pre-trained and fine-tuned features.} The results indicate that features fine-tuned using adapters without ReLU present lower similarity to pre-trained features, in comparison to those fine-tuned with ReLU-based adapters.}
\label{fig:sim}
\vskip -0.2in 
\end{figure}

\textbf{Analysis Results: }
In the Leaky-ReLU experiments, as shown in Figure~\ref{fig:alpha}, performance closely correlates with $\alpha$ values, with outcomes near the extremes resembling those observed at $\alpha=0$ and $\alpha=1$. It is evident that adapters using Leaky-ReLU with multiple non-zero $\alpha$ values outperform those using ReLU ($\alpha = 0$), especially in terms of novel category accuracy. Interestingly, the strong performance observed in the purely linear case ($\alpha = 1$) suggests that the sparsity and non-linearity induced by ReLU may not be essential, and can even be detrimental. In the Threshold-ReLU experiments, as shown in Figure~\ref{fig:t}, performance exhibits a clear dependence on the threshold $t$. Specifically, Threshold-ReLU with negative thresholds ($t \in \{-2, -1\}$) achieves better results than standard ReLU ($t = 0$), whereas positive thresholds ($t \in \{1, 2\}$) lead to a substantial drop in performance. The results in Figure~\ref{fig:t} also illustrate a clear trend from highly sparse settings ($t = 2$) to dense (purely linear case, approximated as $t = -\infty$), further highlighting the negative impact of ReLU-induced sparsity. In the ELU experiments, as shown in Figure~\ref{fig:beta}, enlarging the slope $\beta$ leads to consistent improvements, with all non-zero values ($\beta > 0$) outperforming the standard ReLU ($\beta=0$). These results indicate that enhancing the response to negative values within a reasonable range is conducive to more effective adaptation, whereas merely suppressing negative values to enforce non-linearity does not necessarily yield benefits. Furthermore, Figure~\ref{fig:sim} reveals that adapters without ReLU yield lower feature similarity but better performance, suggesting they enable more effective adaptation compared to those with ReLU.

Overall, the results of the three ReLU variants indicate that, in the context of adapters, the non-linearity introduced by activation functions is not essential for effective adaptation, whereas their inherent sparsity can suppress informative signals, ultimately hindering adaptation.


\begin{figure*}[t]
\begin{center}
\centerline{\includegraphics[width=2.0\columnwidth]{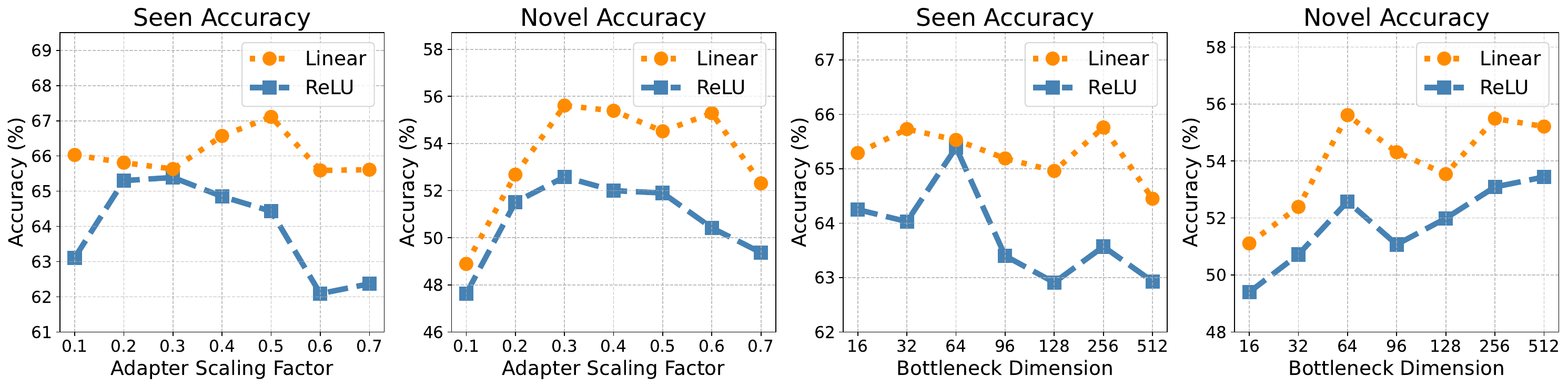}}
\caption{\textbf{Performance comparison between ReluAdapter and LinearAdapter}. Across different adapter scaling factors $\mathit{s}_a$ and bottleneck dimensions $\hat{d}$ on the FGVC-Aircraft, LinearAdapter demonstrates superior overall accuracy.}
\label{fig:adapt_scale}
\end{center}
\vskip -0.25in
\end{figure*}

\subsection{Eliminating the Sparsity of Adapter}

In the above analysis, we observed that mitigating the sparsity induced by ReLU can yield notable performance gains. Although using a fixed $\alpha$ in Leaky-ReLU, a fixed threshold $t$ in Threshold-ReLU or a fixed $\beta$ in ELU can lead to improvements, determining suitable values for these parameters often requires extensive empirical tuning. Our experiments reveal that a simple linear transformation not only exhibits strong generalization capability but also provides superior computational efficiency compared to the non-linear alternatives Leaky-ReLU, Threshold-ReLU, and ELU. Based on these findings, we simply define LinearAdapter as follows:
\begin{equation}
    \label{cal::LinearAdapter}
    \begin{aligned}
        \boldsymbol{\tilde{x}}_\ell = ({\rm LN}(\boldsymbol{x'}_\ell)\cdot\textbf{W}_{\rm down}^{\top} + \textbf{b}_{\rm down})\cdot\textbf{W}_{\rm up}^{\top} + \textbf{b}_{\rm up}
    \end{aligned}
\end{equation}
where $\bm{W}_{\rm down}$ and $\bm{W}_{\rm up}$ represent the down-projection and up-projection matrices with their corresponding bias terms $\textbf{b}_{\rm down}$ and $\textbf{b}_{\rm up}$. The implementation of LinearAdapter, as shown in Figure~\ref{fig:pytorch code}, modifies the adapter by removing ReLU.

We further experiment with different combinations of the bottleneck dimension $\hat{d}$ and the scaling factor $s_a$. As shown in Figure~\ref{fig:adapt_scale}, LinearAdapter shows improvements in both performance and stability. The results also suggest that both small $\hat{d}$ and $s_a$ limit model adaptation, whereas large values lead to training instability.



\begin{figure}[!h]
\begin{python}
class LinearAdapter(nn.Module):
    def forward(self, x):
        x = self.layer_norm(x) 
        x = self.down_proj(x)
        # x = nn.ReLU(x)
        x = self.drop_out(x)
        x = self.up_proj(x)
        x = x * self.s
        return x
\end{python}
\caption{\textbf{A simple modification in PyTorch pseudo-code.}}
\label{fig:pytorch code}
\end{figure}

\textbf{Discussion.} Further analysis reveals that the effects of hyperparameters differ between seen and novel categories. We find that variations cause only minor fluctuations in seen category accuracy, but have a significant impact on novel category accuracy, as shown in Figure~\ref{fig:adapt_scale}. Specifically, increasing $s_a$ or $\hat{d}$ rarely leads to improvements in seen category accuracy; instead, gains in overall accuracy are typically driven by better novel category clustering. This is mainly because the presence of labeled data causes the model to prioritize learning discriminative features from seen categories, which restricts novel category learning when the adapter's capacity is limited. However, setting either parameter too high leads to degraded performance on seen categories, suggesting potential overfitting. In this context, \textbf{LinearAdapter leverages its dense representations to achieve a more favorable trade-off between adaptability and stability than the sparse ReluAdapter}. 

Moreover, other smooth activation functions, such as GeLU and Swish, can also be employed to mitigate feature sparsity, as examined in the subsequent section. Nevertheless, we find that the overall performance is more strongly associated with the output range and gradient saturation of the activation functions, rather than with the specific form of non-linearity. In general, the less compressive the activation function, the better the performance. This observation, to some extent, explains why a simple linear transformation can outperform these non-linear counterparts.

\begin{figure}[t]
\centering
\includegraphics[width=0.95\linewidth]{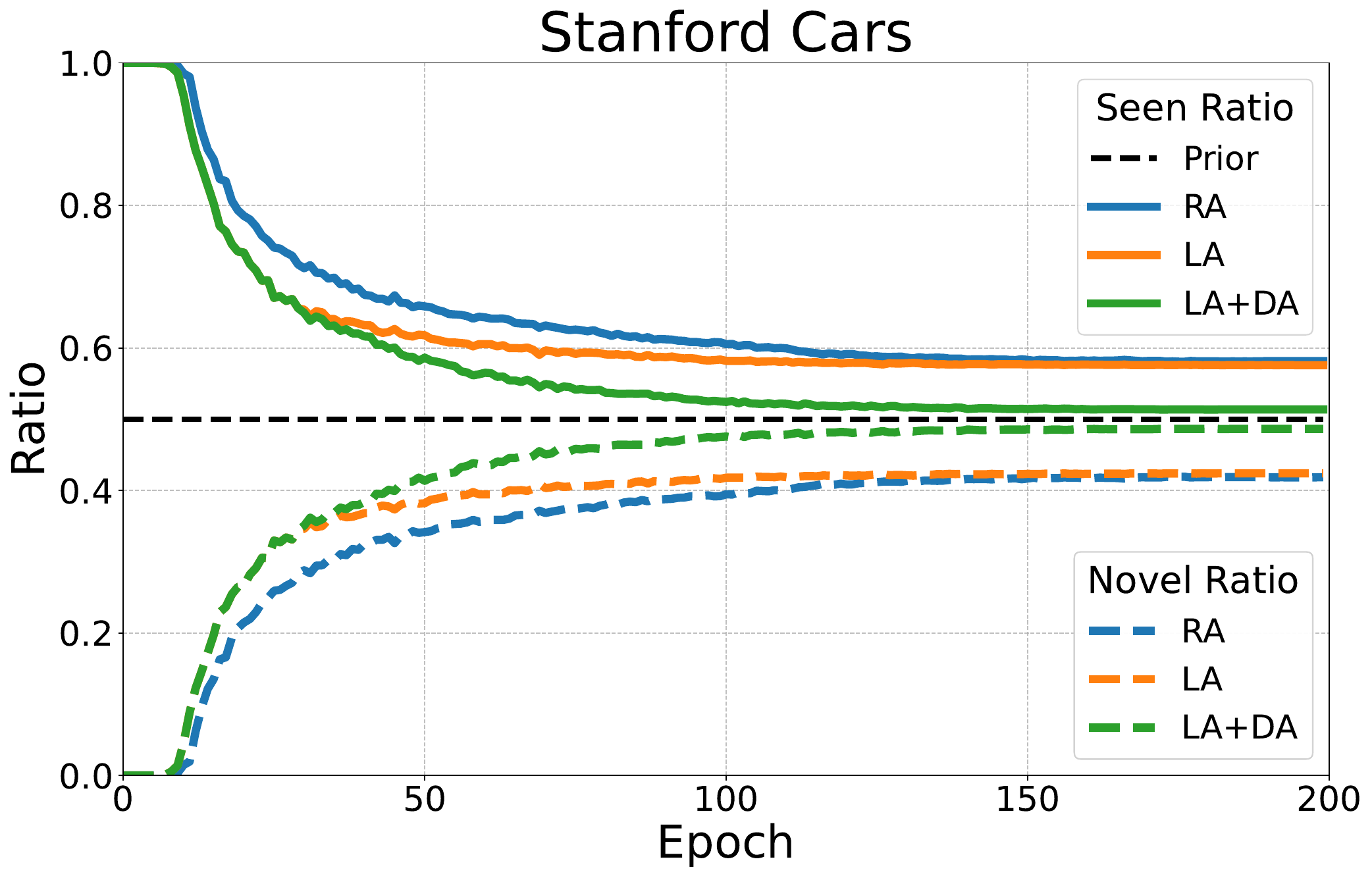}
\caption{\textbf{Fluctuations in predicted bias between seen and novel categories.} We track the change of predictions on the entire dataset to examine how the ratio of samples classified as seen versus novel categories changes across epochs.}
\label{fig:transfer}
\end{figure}

\begin{figure}[t]
\centering
\includegraphics[width=0.95\linewidth]{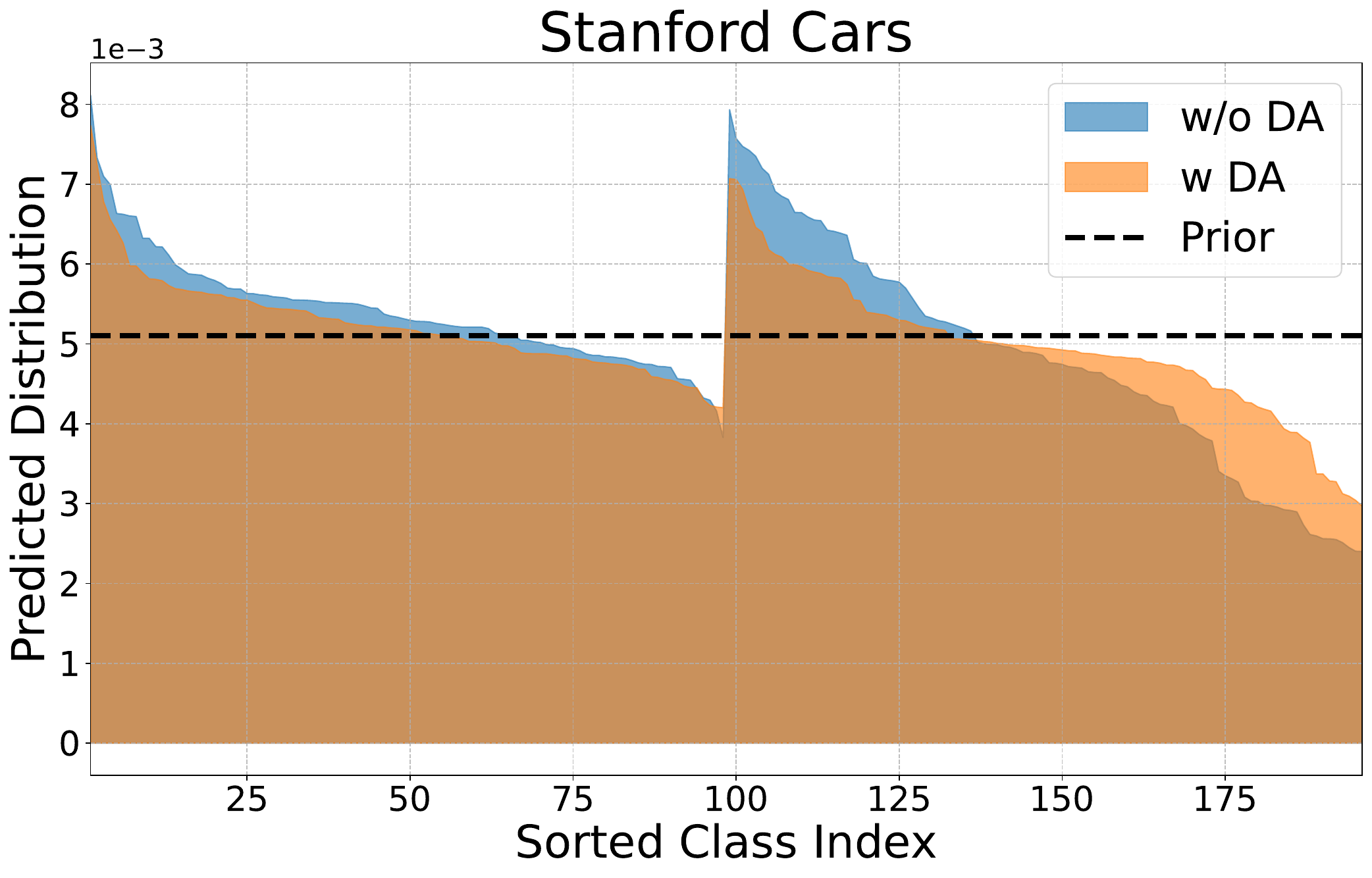}
\caption{\textbf{Comparison of predicted distributions without and with distribution alignment (DA)}. The x-axis indicates sorted class indices, with seen categories in the first half and novel categories in the second half.}
\label{fig:debias_dist}
\end{figure}

\subsection{Debiasing with Distribution Alignment}
GCD methods often suffer from severe bias arising from the asymmetric learning paces between seen and novel categories~\cite{cao2022openworld}. As shown in Figure~\ref{fig:transfer}, when SimGCD is combined with either ReluAdapter (RA) or LinearAdapter (LA), the model still exhibits a clear gap between the predicted and the prior seen sample ratio. This is often caused by an early bias toward seen categories, which could hinder the model's ability to capture key patterns for novel categories. Notably, LinearAdapter shows greater adaptability than ReluAdapter, enabling the model to identify novel classes earlier and ultimately achieve better overall performance. Moreover, an often overlooked detail is that most existing GCD methods rely on a balanced sampler to maintain an equal proportion of labeled and unlabeled samples within each batch, thus achieving a trade-off between supervised and unsupervised contrastive learning. The balanced sampler inevitably oversamples labeled data and undersamples unlabeled data, which in turn accelerates convergence on seen categories, as labeled samples provide accurate cross-entropy supervision. The resulting bias is reflected in the predicted probability distributions: some novel categories receive overconfident predictions, and the distributions of novel categories exhibit higher variance compared to seen categories, as shown in Figure~\ref{fig:debias_dist}.

To address this potential bias without compromising contrastive representation learning, we introduce an auxiliary distribution alignment method. Specifically, we first construct an unaugmented dataset $\mathcal{D}_{\text{v}}$ containing labeled samples from $\mathcal{D}_l$ and unlabeled samples from $\mathcal{D}_u$. We then compute the predicted probability for each sample by traversing the dataset $\mathcal{D}_{\text{v}}$ once, and estimate the overall predicted distribution $\boldsymbol{\pi_v}$ by averaging predictions: $\boldsymbol{\pi_v} = \text{Normalize}(~\sum_{\boldsymbol{x}_i \in \mathcal{D}_{\text{v}}} \boldsymbol{p}_i~)$, where $\boldsymbol{p}_i$ is obtained by Eq.(\ref{eq:output_prob}). Based on the default balanced distribution $\boldsymbol{\pi_b}$, we calculate the distribution alignment vector $\boldsymbol{\pi} = \log \left( {\boldsymbol{\pi_v}}/{\boldsymbol{\pi_b}} \right) \cdot \mathit{s}_d$, where $\mathit{s}_d$ is a scaling factor. Then we modify the unsupervised classification loss $\mathcal{L}^u_\text{cls}$ in Eq.(\ref{eq:selfdistill2}) by replacing its $\ell(\boldsymbol{q}^\prime, \boldsymbol{p})$ with the alignment loss $\ell_{align}(\boldsymbol{q}^{\prime}, \boldsymbol{p})$:
\begin{equation}
\begin{aligned}
\ell_{align}(\boldsymbol{q}^{\prime}, \boldsymbol{p}) = -\sum_{k} {\boldsymbol{q}^{\prime(k)}}\log~(({\boldsymbol{p} +\boldsymbol{\pi}})^{(k)})
\end{aligned}
\end{equation}
Additionally, to enable smoother adjustment and leverage more reliable pseudo-labels, we only apply the alignment loss $\ell_{align}$ after the warm-up phase ends. Interestingly, empirical results indicate that even when the distribution $\boldsymbol{\pi_v}$ is computed \textbf{only once} at the end of the warm-up phase and $\mathit{s}_d$ is set to an appropriate value, which is the default setting in our experiments, the alignment is also effective. As shown in Figure~\ref{fig:transfer} and Figure~\ref{fig:debias_dist}, applying distribution alignment enhances consistency with the prior and results in less biased predictions for novel categories.

\section{Experiment}
\subsection{Experimental Setup}\label{sec:detail1}
\smallskip
\noindent\textbf{Datasets.} We evaluate our method on three generic image recognition datasets: CIFAR-10/100~\cite{krizhevsky2009learning} and ImageNet-100~\cite{tian2020contrastive}, as well as three fine-grained datasets: CUB200~\cite{welinder2010caltech}, Stanford Cars~\cite{krause20133d}, and FGVC-Aircraft~\cite{maji2013fine}, which are part of the Semantic Shift Benchmark (SSB)~\cite{vaze2021open}, and the challenging large-scale imbalanced fine-grained dataset, Herbarium-19~\cite{tan2019herbarium}. For each dataset, we first subsample $|\mathcal{C}_s|$ seen (labeled) categories from the entire set of categories. Following~\cite{vaze2022generalized}, we subsample 80\% of the samples from the seen categories in CIFAR-100 and 50\% of the samples in the other datasets to construct labeled dataset $\mathcal{D}_l$, with the remaining images assigned to unlabeled $\mathcal{D}_u$. The detailed statistics of the datasets are provided in the Table~\ref{table:datasplit}.

\smallskip
\noindent\textbf{Evaluation protocol.} Following previous works, we evaluate our method using clustering accuracy. The accuracy metric compares ground-truth labels $y_i$ with predicted labels $\hat{y}_i$ under optimal label assignment, formally defined as: $\text{accuracy} = \frac{1}{|\mathcal{D}^{\prime}|} \sum_{i=1}^{|\mathcal{D}^{\prime}|} \mathds{1}(y_i = \mathcal{G}(\hat{y}_i))$, where \( \mathcal{G} \) denotes the optimal mapping between predicted and ground-truth labels, and $\mathcal{D}^{\prime}$ denotes either the entire unlabeled data or the novel unlabeled data. Specifically, we evaluate performance across three complementary aspects: (1) seen category accuracy, computed as a standard classification task; (2) novel category accuracy, obtained by solving the assignment between novel-category predictions and their corresponding targets using the Hungarian algorithm~\cite{kuhn55}; and (3) overall accuracy, calculated through optimal assignment across the entire unlabeled dataset.

\begin{table*}[t]   
    \begin{center}
    \caption{Dataset details and training configurations.}
    \label{table:datasplit}
    \resizebox{0.99\linewidth}{!}{  
    \begin{tabular}{llrcrcccccc}  
    \toprule
            &       & \multicolumn{2}{c}{Labeled}  & \multicolumn{2}{c}{Unlabeled}  & \multicolumn{5}{c}{Configs} \\
            \cmidrule(rl){3-4}\cmidrule(rl){5-6}\cmidrule(rl){7-11}  
    Dataset         & Type  & \#Num   & \#Class   & \#Num   & \#Class    & $\texttt{lr}$ & $\lambda_{ent}$ & $\hat{d}$ & $\mathit{s}_a$ &  $\mathit{s}_d$\\
    \midrule
    CIFAR10         & Generic/Balanced & 12.5K     & 5         & 37.5K     & 10  & 0.1  & 1.0 & 64 & 0.1 & 0.2 \\
    CIFAR100        & Generic/Balanced & 20.0K     & 80        & 30.0K     & 100  & 0.2 & 2.0 & 96 & 0.1 & 0.2 \\
    ImageNet100    & Generic/Balanced & 31.9K     & 50        & 95.3K     & 100  & 0.2  & 1.0 & 96 & 0.1 & 0.2 \\
    CUB200             & Fine-grained/Balanced  & 1.5K      & 100       & 4.5K      & 200  & 0.1 & 2.0 & 64 & 0.1 & 0.2 \\
    Stanford Cars   & Fine-grained/Balanced & 2.0K      & 98        & 6.1K      & 196  & 0.1 &  1.0 & 64 & 0.25 & 0.2 \\    
    FGVC-Aircraft   & Fine-grained/Balanced & 1.7K      & 50        & 5.0K      & 50  & 0.1 & 1.0 & 64 & 0.25 & 0.2 \\
    Herbarium19    & Fine-grained/Imbalanced & 8.9K      & 341       & 25.4K     & 683  & 0.2 & 1.0 & 64 & 0.1 & 0.2 \\
    \bottomrule
    \end{tabular}
    }
    \end{center}
\end{table*}

\begin{table*}[t]
\begin{center}
\renewcommand{\arraystretch}{0.8}
\setlength{\tabcolsep}{0.5pt}
\caption{{Experimental results on the fine-grained benchmark}. Our results are obtained from three runs with fixed random seeds (0, 1, 2) for reproducibility, and we report mean and standard deviation. Bold and underlined values indicate the best and second-best results, respectively.}
\label{table:ssb}
\scalebox{1.06}{
\begin{tabular}{lccccccccccccc}
\hline\noalign{\smallskip}
\multirow{2}{*}{\textbf{Methods}} & \multicolumn{3}{c}{CUB200} & \multicolumn{3}{c}{Stanford Cars} & \multicolumn{3}{c}{FGVC-Aircraft} & \multicolumn{3}{c}{Herbarium19} & \multirow{2}{*}{\textbf{Avg.}} \\
\cmidrule(lr){2-4}\cmidrule(lr){5-7}\cmidrule(lr){8-10}\cmidrule(lr){11-13}
 & \cellcolor{blue!10}All & Seen & Novel & \cellcolor{blue!10}All & Seen & Novel & \cellcolor{blue!10}All & Seen & Novel  & \cellcolor{blue!10}All & Seen & Novel \\
\noalign{\smallskip}
\hline
\noalign{\smallskip}
$k\mbox{-}means$~\cite{arthur2006k}   &\cellcolor{blue!10}34.3   & 38.9& 32.1& \cellcolor{blue!10}12.8& 10.6& 13.8& \cellcolor{blue!10}16.0& 14.4& 16.8 & \cellcolor{blue!10}13.0 & 12.2 & 13.4 & \cellcolor{blue!10}19.0 \\
RankStats+~\cite{han2021autonovel}    &\cellcolor{blue!10}33.3 &51.6 &24.2 &\cellcolor{blue!10}28.3 &61.8 &12.1& \cellcolor{blue!10}26.9& 36.4& 22.2 & \cellcolor{blue!10}27.9 & 55.8 & 12.8 & \cellcolor{blue!10}32.8\\
UNO+~\cite{fini2021unified}           &\cellcolor{blue!10}35.1 &49.0 &28.1 &\cellcolor{blue!10}35.5 &70.5 &18.6& \cellcolor{blue!10}40.3& 56.4& 32.2 & \cellcolor{blue!10}28.3 & 53.7 & 14.7 & \cellcolor{blue!10}38.5\\
\noalign{\smallskip}
\hline
\noalign{\smallskip}
ORCA~\cite{cao2022openworld}                &\cellcolor{blue!10}35.3 &45.6 &30.2 &\cellcolor{blue!10}23.5 &50.1 &10.7& \cellcolor{blue!10}22.0& 31.8& 17.1 & \cellcolor{blue!10}35.4 & 51.0 & 27.0 & \cellcolor{blue!10}31.6\\
GCD~\cite{vaze2022generalized}             &\cellcolor{blue!10}51.3 &56.6 &48.7 &\cellcolor{blue!10}39.0 &57.6 &29.9& \cellcolor{blue!10}45.0& 41.1& 46.9 & \cellcolor{blue!10}20.9 & 30.9 & 15.5 & \cellcolor{blue!10}40.3\\
XCon~\cite{fei2022xcon}             &\cellcolor{blue!10}52.1 &54.3 &51.0 &\cellcolor{blue!10}40.5 &58.8 &31.7& \cellcolor{blue!10}47.7& 44.4& 49.4 & \cellcolor{blue!10}- & - & - & \cellcolor{blue!10}-\\
DCCL~\cite{pu2023dynamic}                 &\cellcolor{blue!10}63.5 &60.8 &64.9 &\cellcolor{blue!10}43.1 &55.7 &36.2& \cellcolor{blue!10}- & - & - & \cellcolor{blue!10}- & - & - & \cellcolor{blue!10}-\\
GPC~\cite{zhao2023learning}         &\cellcolor{blue!10}55.4 &58.2 &53.1 &\cellcolor{blue!10}42.8 &59.2 &32.8& \cellcolor{blue!10}46.3& 42.5& 47.9 & \cellcolor{blue!10}- & - & - & \cellcolor{blue!10}-\\
SimGCD~\cite{wen2023parametric}         &\cellcolor{blue!10}60.3 &65.6 &57.7 &\cellcolor{blue!10}53.8 &71.9 &45.0& \cellcolor{blue!10}54.2& 59.1& 51.8 & \cellcolor{blue!10}44.0 & 58.0 & 36.4 & \cellcolor{blue!10}54.8\\ 
PromptCAL~\cite{zhang2023promptcal}  &\cellcolor{blue!10}62.9 &64.4 &62.1 &\cellcolor{blue!10}50.2 &70.1 &40.6& \cellcolor{blue!10}52.2& 52.2& 52.3 & \cellcolor{blue!10}37.0 & 52.0 & 28.9  & \cellcolor{blue!10}52.1\\
AMEND~\cite{banerjee2024amend}   &\cellcolor{blue!10}64.9 &\underline{75.6} &59.6&\cellcolor{blue!10}56.4 &73.3 &48.2 &\cellcolor{blue!10}52.8 &61.8 & 48.3 &\cellcolor{blue!10}44.2 &60.5 &35.4 & \cellcolor{blue!10}56.8\\
$\mu$GCD~\cite{vaze2024no}   &\cellcolor{blue!10}65.7&68.0 &64.6&\cellcolor{blue!10}56.5 &68.1 &50.9 &\cellcolor{blue!10}53.8 &55.4 & 53.0 &\cellcolor{blue!10}\underline{45.8} &\underline{61.9} &37.2 &\cellcolor{blue!10}56.7\\
SPTNet~\cite{wang2024sptnet}                               &\cellcolor{blue!10}65.8&68.8 &65.1 &\cellcolor{blue!10}59.0 &79.2 &49.3 &\cellcolor{blue!10}\underline{59.3}&61.8& \textbf{58.1} &\cellcolor{blue!10}{43.4} & 58.7 & {35.2} & \cellcolor{blue!10}58.7\\
CMS~\cite{choi2024contrastive}                               &\cellcolor{blue!10}68.2 &\textbf{76.5} &64.0 &\cellcolor{blue!10}56.9 &76.1 &47.6 &\cellcolor{blue!10}56.0& 63.4& 52.3 &\cellcolor{blue!10}{36.4} & 54.9 & {26.4} & \cellcolor{blue!10}56.6\\
LegoGCD~\cite{cao2024solving}                              &\cellcolor{blue!10}63.8 &71.9 &59.8 &\cellcolor{blue!10}57.3 &75.7 &48.4 &\cellcolor{blue!10}55.0& 61.5& 51.7 &\cellcolor{blue!10}45.1 & 57.4 & \underline{38.4} & \cellcolor{blue!10}57.2\\
ProtoGCD~\cite{ma2025protogcd}         &\cellcolor{blue!10}63.2 &68.5 &60.5 &\cellcolor{blue!10}53.8 &73.7 &44.2& \cellcolor{blue!10}56.8& 62.5& 53.9 & \cellcolor{blue!10}44.5 & 59.4 & 36.5 & \cellcolor{blue!10}56.5\\
AdaptGCD~\cite{qu2024adaptgcd}                              &\cellcolor{blue!10}\underline{68.8} &74.5 &\underline{65.9} &\cellcolor{blue!10}\underline{62.7} &\textbf{80.6} &\underline{54.0} &\cellcolor{blue!10}57.9& \underline{65.2}& 54.3 &\cellcolor{blue!10}45.7 & 60.6 & 37.7 & \cellcolor{blue!10}\underline{60.7}\\

\noalign{\smallskip}
\hline
\noalign{\smallskip}
\textbf{LAGCD (ours)} &\cellcolor{blue!10}\textbf{69.1{\scriptsize $\pm$1.6}} &75.2{\scriptsize $\pm$0.9}&\textbf{66.1{\scriptsize $\pm$2.8}}&\cellcolor{blue!10}\textbf{63.6{\scriptsize $\pm$1.2}}&\underline{79.6{\scriptsize $\pm$1.6}}&\textbf{55.9{\scriptsize $\pm$1.2}}&\cellcolor{blue!10}\textbf{59.6{\scriptsize $\pm$1.0}}&\textbf{67.4{\scriptsize $\pm$0.9}}&\underline{55.7{\scriptsize $\pm$1.4}}&\cellcolor{blue!10}\textbf{49.3{\scriptsize $\pm$0.4}}&\textbf{67.2{\scriptsize $\pm$0.1}}&\textbf{39.6{\scriptsize $\pm$0.7}}& \cellcolor{blue!10}\textbf{62.4}\\

\hline
\end{tabular}
}
\end{center}
\vskip -0.1in
\end{table*}

\smallskip
\noindent\textbf{Implementation details.} Following previous works, we adopt a DINO-pretrained ViT-B/16~\cite{caron2021emerging, dosovitskiy2020image} as our backbone, followed by a projection head for contrastive representation learning and a parametric prototype classifier. We take the feature of the $\texttt{CLS}$ token from the backbone as the image feature. And we employ \textbf{a batch size of 128 over 200 epochs}, with an initial learning rate of $\texttt{lr}$ that follows a cosine decay schedule. Following SimGCD~\cite{wen2023parametric}, we set the supervised balancing factor $\lambda_{sup}=0.35$, temperature parameters $\tau_u=1.0$, and $\tau_c=0.07$ for $\mathcal{L}^u_\text{rep}$ and $\mathcal{L}^s_\text{rep}$, respectively. For the classification objective, we employ a temperature parameter $\tau_s=0.1$. The temperature $\tau_t$ is initialized as 0.07 and gradually annealed to 0.04 over the first 30 epochs following a cosine schedule. Unlike previous methods, we integrate adapters into each ViT block with a bottleneck dimension of $\hat{d}$, and \textbf{keep all backbone parameters frozen}. Detailed hyperparameters, including $\texttt{lr}$, $\lambda_{ent}$, $\hat{d}$, $\mathit{s}_a$, and $\mathit{s}_d$, are provided in the Table~\ref{table:datasplit}. Experimental results in the Table~\ref{table:ssb} and Table~\ref{table:generic} are conducted using a single NVIDIA Tesla V100 GPU.

In the Table~\ref{table:datasplit}, we provide the detailed information about the datasets and hyperparameters. For datasets, we utilize seven commonly used datasets in the GCD task, including four fine-grained datasets and three generic datasets. Moreover, we only introduce two necessary hyperparameters ($\hat{d}$ and $\mathit{s}_a$) for adapter-based tuning, as well as the scaling factor $\mathit{s}_d$ for distribution alignment, and most of them are consistent without much tuning.

\subsection{Comparison with Existing Methods}
\textbf{Baselines.} We conduct comprehensive comparison against multiple SOTA GCD methods, including AdaptGCD~\cite{qu2024adaptgcd}, ProtoGCD~\cite{ma2025protogcd}, LegoGCD~\cite{cao2024solving}, CMS~\cite{choi2024contrastive}, SPTNet~\cite{wang2024sptnet}, AMEND~\cite{banerjee2024amend},
$\mu$GCD~\cite{vaze2024no},
PromptCAL~\cite{zhang2023promptcal},
DCCL~\cite{pu2023dynamic},
GPC~\cite{zhao2023learning},
SimGCD~\cite{wen2023parametric}, XCon~\cite{fei2022xcon}, GCD~\cite{vaze2022generalized} and ORCA~\cite{cao2022openworld}. We compare ours against NCD baselines RankStats+~\cite{han2021autonovel} and UNO+~\cite{fini2021unified}, and also compare against the $k$-means clustering~\cite{arthur2006k} applied to features extracted from the pre-trained DINO~\cite{caron2021emerging}.

\begin{table*}[t]
\begin{center}
\renewcommand{\arraystretch}{0.75}
\setlength{\tabcolsep}{2pt}
\caption{Experimental results on the generic benchmark. Our results are obtained from three runs with fixed random seeds (0, 1, 2) for reproducibility, and we report mean and standard deviation. Bold and underlined values indicate the best and second-best results, respectively.}
\label{table:generic}
\scalebox{1.1}{
\begin{tabular}{lcccccccccc}
\hline\noalign{\smallskip}
\multirow{2}{*}{\textbf{Methods}}& \multicolumn{3}{c}{CIFAR10} & \multicolumn{3}{c}{CIFAR100} & \multicolumn{3}{c}{ImageNet100} & \multirow{2}{*}{\textbf{Avg.}}\\
\cmidrule(lr){2-4}\cmidrule(lr){5-7}\cmidrule(lr){8-10}
 & \cellcolor{blue!10} All & Seen & Novel & \cellcolor{blue!10} All & Seen & Novel & \cellcolor{blue!10} All & Seen & Novel \\
\noalign{\smallskip}
\hline
\noalign{\smallskip}
$k$-means~\cite{arthur2006k}    & \cellcolor{blue!10} 83.6& 85.7& 82.5& \cellcolor{blue!10} 52.0& 52.2& 50.8& \cellcolor{blue!10} 72.7& 75.5& 71.3 & \cellcolor{blue!10}69.6\\
RankStats+~\cite{han2021autonovel}      & \cellcolor{blue!10} 46.8& 19.2& 60.5& \cellcolor{blue!10} 58.2& 77.6& 19.3& \cellcolor{blue!10} 37.1& 61.6& 24.8& \cellcolor{blue!10}45.0\\
UNO+~\cite{fini2021unified}              & \cellcolor{blue!10} 68.6& 98.3& 53.8& \cellcolor{blue!10} 69.5& 80.6& 47.2& \cellcolor{blue!10} 70.3& 95.0& 57.9& \cellcolor{blue!10}71.2\\
\noalign{\smallskip}
\hline
\noalign{\smallskip}
ORCA~\cite{cao2022openworld}                  & \cellcolor{blue!10} 81.8& 86.2& 79.6& \cellcolor{blue!10} 69.0& 77.4& 52.0& \cellcolor{blue!10} 73.5& 92.6& 63.9& \cellcolor{blue!10}75.1\\
GCD~\cite{vaze2022generalized}              & \cellcolor{blue!10} 91.5& \underline{97.9}& 88.2& \cellcolor{blue!10} 73.0& 76.2& 66.5& \cellcolor{blue!10} 74.1& 89.8& 66.3& \cellcolor{blue!10}80.3\\
XCon~\cite{fei2022xcon}              & \cellcolor{blue!10} 96.0& 97.3& 95.4& \cellcolor{blue!10} 74.2& 81.2& 60.3& \cellcolor{blue!10} 77.6& 93.5& 69.7& \cellcolor{blue!10}82.8\\
DCCL~\cite{pu2023dynamic}                   & \cellcolor{blue!10} 96.3& 96.5& 96.9& \cellcolor{blue!10} 75.3& 76.8& 70.2& \cellcolor{blue!10} 80.5& 90.5& 76.2& \cellcolor{blue!10}87.2\\
GPC~\cite{zhao2023learning} & \cellcolor{blue!10} 92.2& \textbf{98.2}& 89.1& \cellcolor{blue!10} 77.9& 85.0& 63.0& \cellcolor{blue!10} 76.9& 94.3& 71.0& \cellcolor{blue!10}83.1\\
SimGCD~\cite{wen2023parametric}          & \cellcolor{blue!10} 97.1& 95.1& 98.1& \cellcolor{blue!10} 80.1& 81.2& 77.8& \cellcolor{blue!10} 83.0& 93.1& 77.9& \cellcolor{blue!10}87.0\\
PromptCAL~\cite{zhang2023promptcal} & \cellcolor{blue!10} \underline{97.9}& 96.6& 98.5& \cellcolor{blue!10} 81.2& 84.2& 75.3& \cellcolor{blue!10} 83.1& 92.7& 78.3& \cellcolor{blue!10}87.5\\
AMEND~\cite{banerjee2024amend} & \cellcolor{blue!10} 96.8& 94.6& 97.8& \cellcolor{blue!10} 81.0& 79.9& \underline{83.3}& \cellcolor{blue!10} 83.2& 92.9& 78.3& \cellcolor{blue!10}87.5\\
SPTNet~\cite{wang2024sptnet}    & \cellcolor{blue!10} 97.3& 95.0 &98.6 & \cellcolor{blue!10} 81.3 &84.3 &75.6 & \cellcolor{blue!10} 85.4 &93.2 &\underline{81.4}& \cellcolor{blue!10}88.0\\
CMS~\cite{choi2024contrastive}                  & \cellcolor{blue!10}- & - & - & \cellcolor{blue!10} 82.3& 85.7& 75.5& \cellcolor{blue!10} 84.7& \textbf{95.6}& 79.2& \cellcolor{blue!10}-\\
LegoGCD~\cite{cao2024solving}          & \cellcolor{blue!10} 97.1& 94.3& 98.5& \cellcolor{blue!10} 81.8& 81.4& 82.5& \cellcolor{blue!10} \underline{86.3}& 94.5& \textbf{82.1}& \cellcolor{blue!10}88.7\\
ProtoGCD~\cite{ma2025protogcd}          & \cellcolor{blue!10} 97.3& 95.3& 98.2& \cellcolor{blue!10} 81.9& 82.9& 80.0& \cellcolor{blue!10} 84.0& 92.2& 79.9& \cellcolor{blue!10}88.0\\
AdaptGCD~\cite{qu2024adaptgcd}          & \cellcolor{blue!10} \underline{97.9}& 95.5& \underline{99.0}& \cellcolor{blue!10} \underline{84.0}& \underline{86.2}& 79.7& \cellcolor{blue!10} \textbf{86.4}& \underline{94.9}& \textbf{82.1}& \cellcolor{blue!10}\underline{89.5}\\
\noalign{\smallskip}
\hline
\noalign{\smallskip}
\textbf{LAGCD (ours)} &  \cellcolor{blue!10}\textbf{98.6{\scriptsize $\pm$0.0}}&97.6{\scriptsize $\pm$0.2}&\textbf{99.1{\scriptsize $\pm$0.1}}& \cellcolor{blue!10}\textbf{86.3{\scriptsize $\pm$1.1}}&\textbf{87.8{\scriptsize $\pm$0.5}}&\textbf{83.5{\scriptsize $\pm$3.4}}&
\cellcolor{blue!10}85.9{\scriptsize $\pm$0.9} & \underline{94.9{\scriptsize $\pm$0.1}} & \underline{81.4{\scriptsize $\pm$1.3}}& \cellcolor{blue!10}\textbf{90.6}\\

\noalign{\smallskip}
\hline
\end{tabular}
}
\end{center}
\end{table*}

\begin{table*}[!h]
    \begin{center}
    \setlength{\tabcolsep}{4.4pt}
    \caption{Ablation study on CUB200, Stanford Cars and Herbarium19.}
    \resizebox{0.8\linewidth}{!}{
    \begin{tabular}{ccccccccccccc}
    \toprule
         &  &  &  &\multicolumn{3}{c}{CUB200} &\multicolumn{3}{c}{Stanford Cars} &\multicolumn{3}{c}{Herbarium19} \\  \cmidrule(lr){5-7} \cmidrule(lr){8-10} \cmidrule(lr){11-13}
       \multirow{-2}{*}{SimGCD} &\multirow{-2}{*}{RA}  &\multirow{-2}{*}{LA} &\multirow{-2}{*}{DA}  &\cellcolor{blue!10}All  &Seen   &Novel &\cellcolor{blue!10}All  &Seen   &Novel &\cellcolor{blue!10}All  &Seen   &Novel\\
       \noalign{\smallskip}
       \hline
       \noalign{\smallskip}
        \ding{51} &  &  & &\cellcolor{blue!10}60.3 &65.6 &57.7 &\cellcolor{blue!10}53.8 &71.9 &45.0 &\cellcolor{blue!10}44.0 &58.0 &36.4\\
        \ding{51} & \ding{51}  &  & &\cellcolor{blue!10}67.5 &73.9 &64.3 &\cellcolor{blue!10}58.7 &77.0 &49.9 &\cellcolor{blue!10}48.0 &66.7 &38.0\\
        \ding{51} &  & \ding{51} & &\cellcolor{blue!10}68.8 &74.9 &65.7 &\cellcolor{blue!10}62.1 &78.5 &54.2 &\cellcolor{blue!10}48.8 &66.7 &39.2\\
        \ding{51} &  & \ding{51} & \ding{51} &\cellcolor{blue!10}\textbf{69.1} &\textbf{75.2} &\textbf{66.1} &\cellcolor{blue!10}\textbf{63.6} &\textbf{79.6} &\textbf{55.9} &\cellcolor{blue!10}\textbf{49.3} &\textbf{67.2} &\textbf{39.6}\\
    \bottomrule
    \end{tabular}}
    \label{table:ablation}
    \end{center}
\vskip -0.1in
\end{table*}

\smallskip
\noindent\textbf{Results on fine-grained datasets.} 
Table~\ref{table:ssb} presents results on four fine-grained datasets. Fine-grained category discovery poses a greater challenge compared to generic category discovery. This enhanced complexity manifests through multiple factors: fine-grained datasets exhibit larger intra-class and smaller inter-class variations, while containing more categories but fewer samples per category. \textbf{Compared to the baseline SimGCD}, our method achieves significant performance improvements, with an average accuracy gain of 7.6\%. \textbf{Compared to SPTNet}, the state-of-the-art method utilizing visual prompts, our method achieves superior performance, surpassing SPTNet by 0.3\% on FGVC-Aircraft, 3.3\% on CUB-200, 4.6\% on Stanford Cars, and 5.9\% on Herbarium19. \textbf{Compared to GCD methods with partial fine-tuning}, our method achieves at least a 5.2\% improvement. 
\textbf{Compared to AdaptGCD}, our method stands out for architectural simplicity and implementation efficiency. While AdaptGCD employs multi-expert adapters with routing constraints to separate seen and novel class processing, requiring multiple adapters per ViT block and additional loss terms for expert balancing, our method imropves adaptation by simply removing ReLU activation in adapters, creating a streamlined linear transformation without added complexity. Our method achieves superior performance with a simpler modification, attaining a higher average accuracy of 62.4\% compared to 60.7\%.

\smallskip
\noindent\textbf{Results on generic datasets.} Table~\ref{table:generic} presents results on three generic datasets. Overall, these datasets exhibit higher average accuracy than fine-grained ones, making further improvements more challenging. Nevertheless, our method achieves SOTA performance by improving the accuracy on CIFAR10 from the previous best of 97.9\% (PromptCAL and AdaptGCD) to 98.6\%, and on CIFAR100 from 84.0\% (AdaptGCD) to 86.3\%. Notably, our method demonstrates a 1.1\% improvement in average accuracy across all three datasets compared to AdaptGCD.

\smallskip
\noindent\textbf{Tunable parameters.} Our method is an end-to-end one-stage adapter-based method that no longer fine-tunes the last block of the backbone. The total number of parameters introduced to pre-trained model is $(d\times \hat{d}\times2  + \hat{d} + d + d\times2)\times n $, where $d$ and $\hat{d}$ denote the input and bottleneck dimension, and $n$ denotes the number of adapted blocks. We insert adapters into all 12 ViT blocks and set $\hat{d}$ to two values: 96 and 64, introducing 1.8M and 1.2M tunable parameters, respectively. Most GCD methods exclusively fine-tune the last block of the backbone, which contains 7.1M parameters. While prompt-based methods (SPTNet and PromptCAL) utilize lightweight visual prompts, they still rely on fine-tuning the last block like conventional GCD methods to achieve better adaptation. And the adapter-based method AdaptGCD requires 6.4M and 4.8M parameters due to its multi-expert adapter structure.

\subsection{Ablation Study}

To evaluate our method LAGCD, we conduct ablation studies on three datasets: CUB200, Stanford Cars, and Herbarium19. All results are averaged over three runs. We analyze the contributions of the baseline SimGCD and components: ReluAdapter (RA), LinearAdapter (LA), and distribution alignment (DA). Table~\ref{table:ablation} presents the detailed results. The baseline SimGCD, which adopts the partial fine-tuning, achieves overall accuracies of 60.3\%, 53.8\%, and 44.0\% on the three datasets respectively. Incorporating ReluAdapter significantly improves performance, increasing overall accuracy by 7.2\%, 4.9\%, and 4.0\%, demonstrating the effectiveness of adapter-based tuning. By removing the intrinsic ReLU activation, LinearAdapter further enhances performance to 68.8\% (+1.3\%), 62.1\% (+3.4\%), and 48.8\% (+0.8\%). The improvement indicates that the non-linearity and sparsity introduced by ReLU within the adapters are not essential and may even hinder overall performance. Moreover, integrating LA with DA consistently yields performance gains, most notably on the Stanford Cars dataset, where overall accuracy improves by 1.5\%.

\begin{figure*}[t]
    \centering
    \begin{minipage}[b]{0.49\textwidth}
        \centering
        \includegraphics[width=\textwidth]{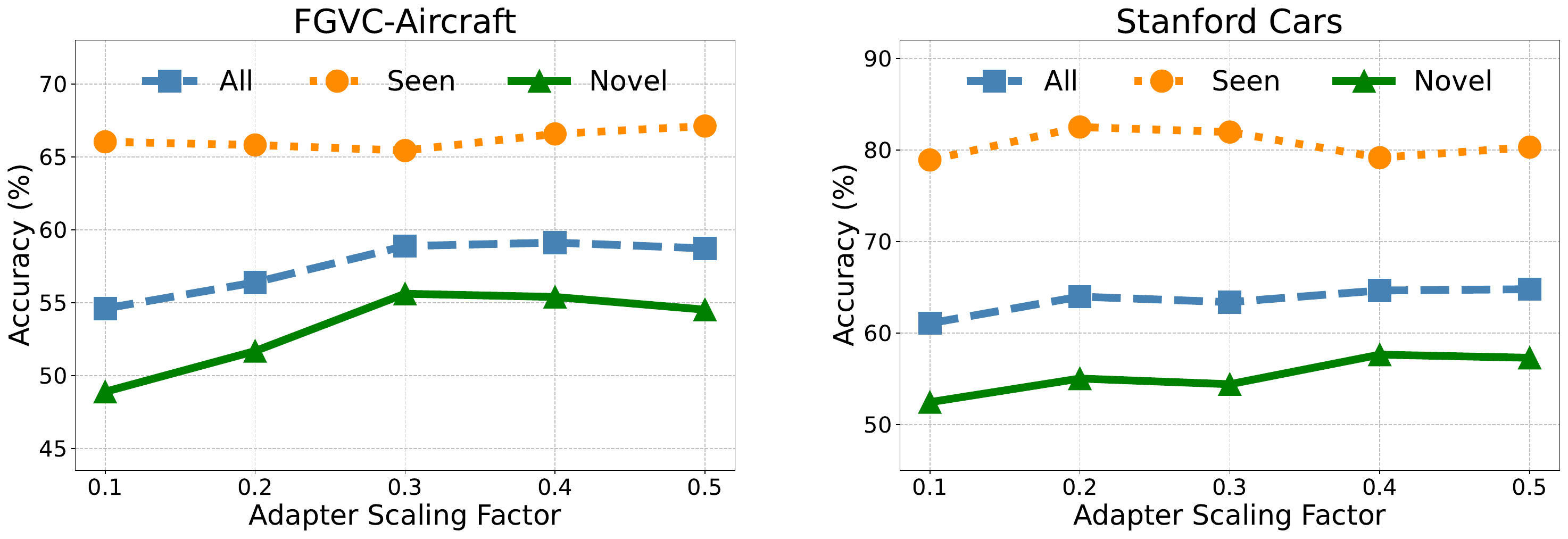}
        \caption{Sensitivity of the scale $\mathit{s}_a$ of adapter.}
        \label{fig:sens_adapter_scale}
    \end{minipage}
    \hfill
    \begin{minipage}[b]{0.49\textwidth}
        \centering
        \includegraphics[width=\textwidth]{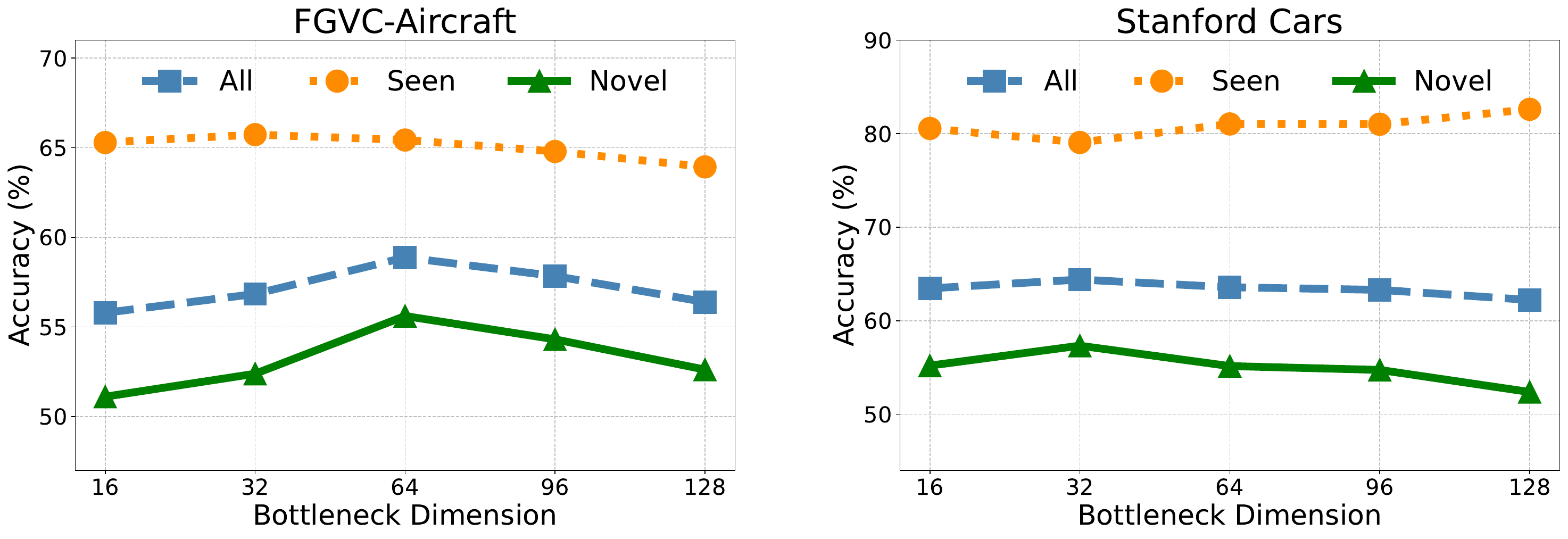}
        \caption{Sensitivity of the bottleneck dimension $\hat{d}$.}
        \label{fig:sens_adapter_dim}
    \end{minipage}
\end{figure*}

\begin{figure*}[t]
    \centering
    \begin{minipage}[b]{0.49\textwidth}
        \centering
        \includegraphics[width=\textwidth]{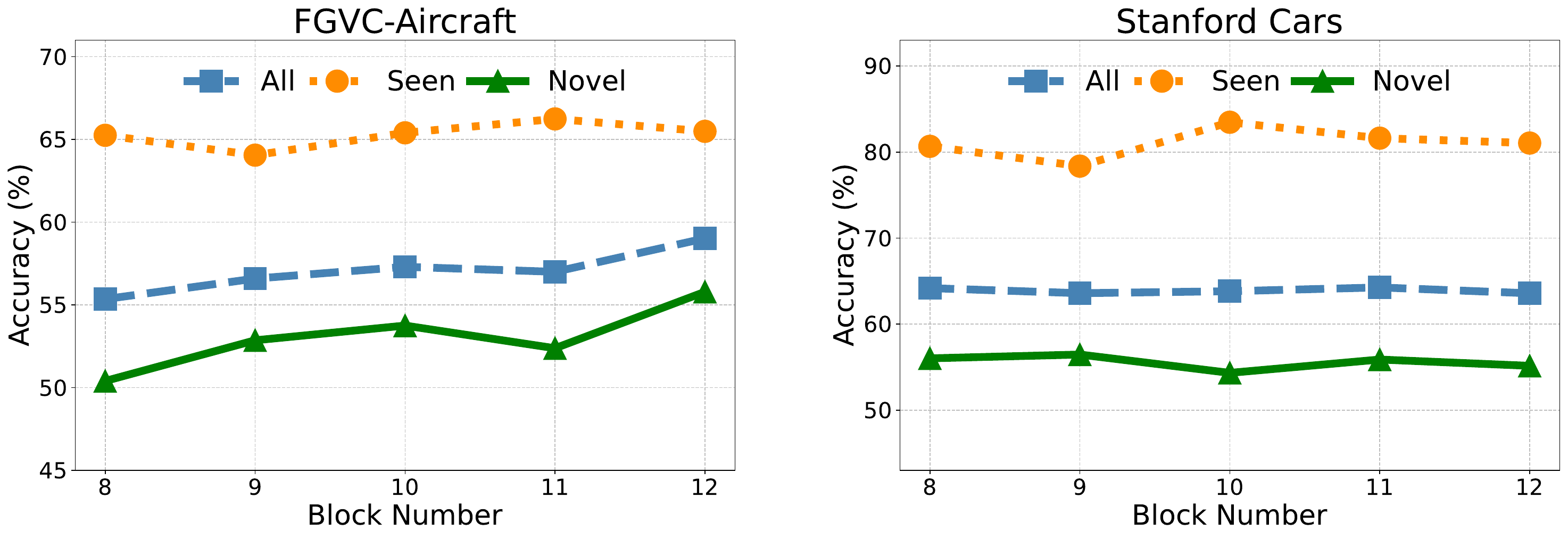}
        \caption{Sensitivity of the number $n$ of adapted blocks.}
        \label{fig:sens_block_num}
    \end{minipage}
    \hfill
    \begin{minipage}[b]{0.49\textwidth}
        \centering
        \includegraphics[width=\textwidth]{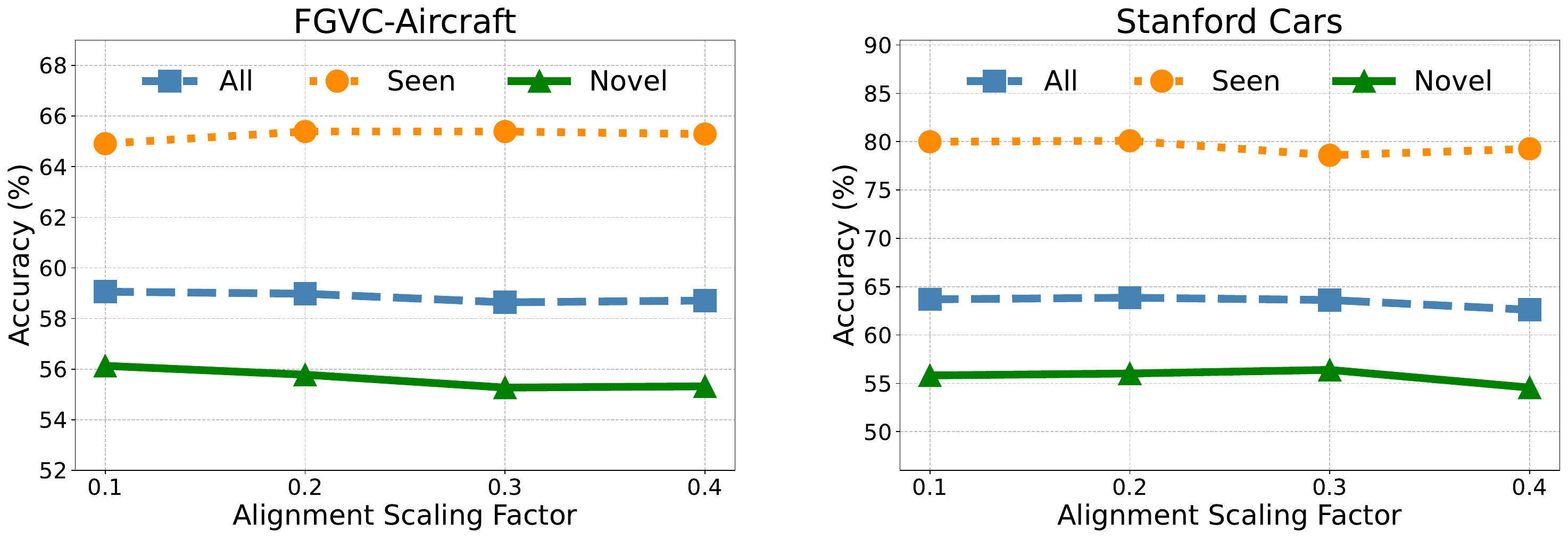}
        \caption{Sensitivity of the alignment scaling factor $\mathit{s}_d$.}
        \label{fig:sens_align_scale}
    \end{minipage}
\end{figure*}

\subsection{Sensitivity Study}

For the sensitivity study, we investigate four key components of LAGCD. Regarding adapters, we evaluate the impact of three hyperparameters: the scaling factor $\mathit{s}_a$, the bottleneck dimension $\hat{d}$, and the number $n$ of adapted blocks. For distribution alignment, we focus on its scaling factor $\mathit{s}_d$. These experiments are primarily conducted on the FGVC-Aircraft and Stanford Cars datasets.

\subsubsection{The scaling factor of adapter}
The scaling factor $\mathit{s}_a$ is utilized to balance features between two branches: task-agnostic features from the frozen original branch and task-related features from the tunable bottleneck branch. As illustrated in Figure~\ref{fig:sens_adapter_scale}, when the scaling factor $\mathit{s}_a$ is too small, the model tends to underfit, resulting in suboptimal performance. As $\mathit{s}_a$ increases to an appropriate value, such as around 0.3, the performance improves significantly. However, when $\mathit{s}_a$ becomes excessively large, the model suffers from overfitting, leading to a decline in overall performance.

\subsubsection{The bottleneck dimension of adapter}
The bottleneck dimension $\hat{d}$ serves as a key hyperparameter that controls the trade-off between computational efficiency and effective adaptation. Our systematic analysis in the previous section shows that smaller bottleneck dimensions, while reducing the parameter count, often degrade performance, whereas larger dimensions increase capacity but may introduce overfitting and training instability. As illustrated in Figure~\ref{fig:sens_adapter_dim}, results reveal a consistent trend across both datasets: accuracy improves as $\hat{d}$ increases from 16 to 32, plateaus between 32 and 128, and then shows no further gain. This indicates that a moderate bottleneck dimension such as around 64 achieves the trade-off between efficiency and accuracy. Importantly, variations in $\hat{d}$ primarily affect novel category accuracy, while performance on seen categories remains relatively stable.

\subsubsection{The number of adapted blocks}

Since adapters have been shown to be more effective in the upper blocks of ViT (those farther from the input image)\cite{chen2022adaptformer}, we systematically examine model performance by varying the number of adapted blocks from top to bottom. As illustrated in Figure\ref{fig:sens_block_num}, increasing the number $n$ of adapted blocks consistently enhances adaptability across datasets. Notably, the results indicate that adapting as few as eight blocks is sufficient to achieve competitive performance. This finding suggests a practical threshold that balances effective adaptation with computational efficiency.

\begin{figure*}[t]
\centering
\begin{minipage}[b]{0.49\linewidth}
\centering
\includegraphics[width=\linewidth]{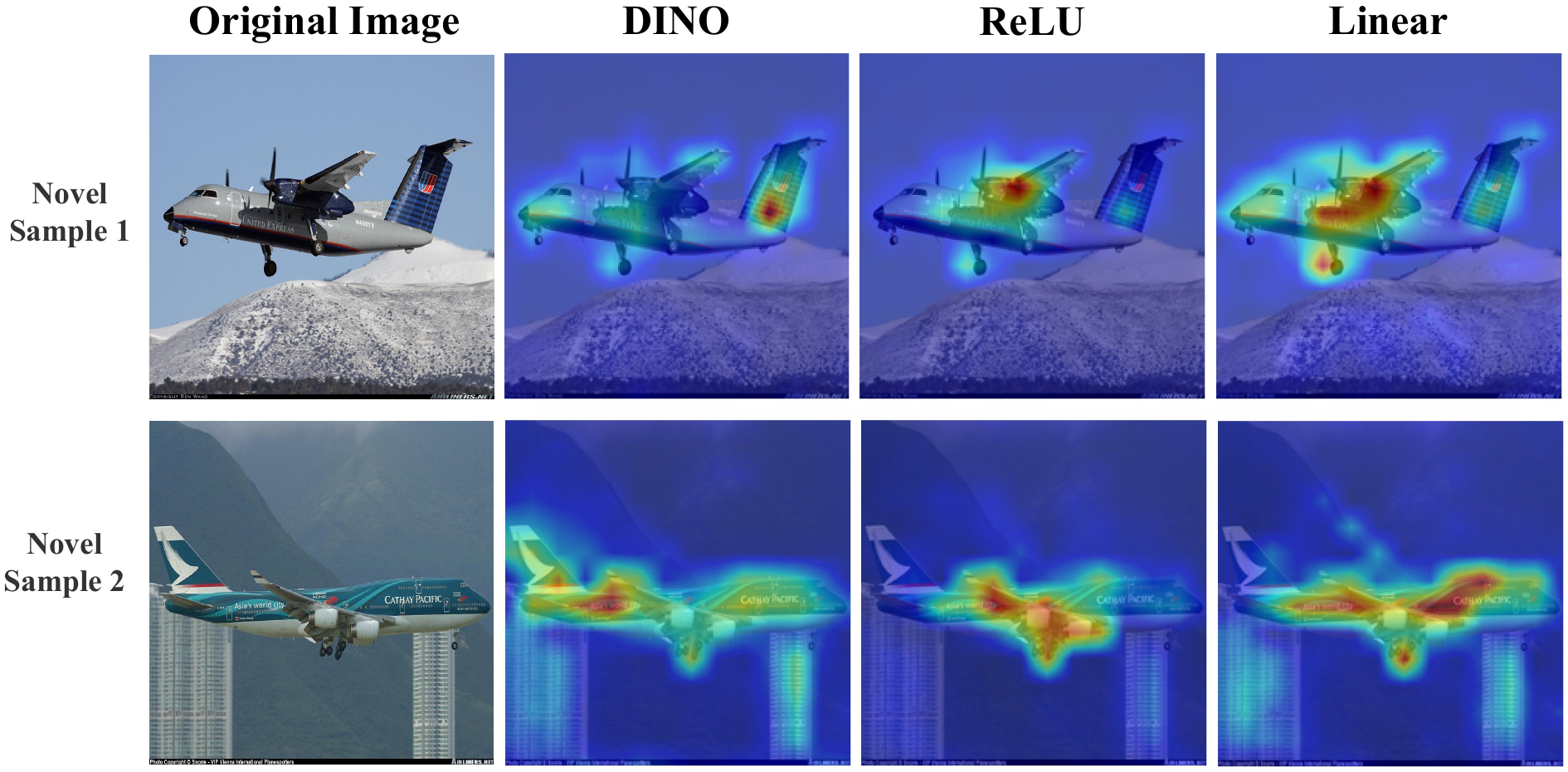}
\caption{\textbf{FGVC-Aircraft Novel Categories.}}
\label{fig:comparison1}
\end{minipage}
\hfill
\begin{minipage}[b]{0.49\linewidth}
\centering
\includegraphics[width=\linewidth]{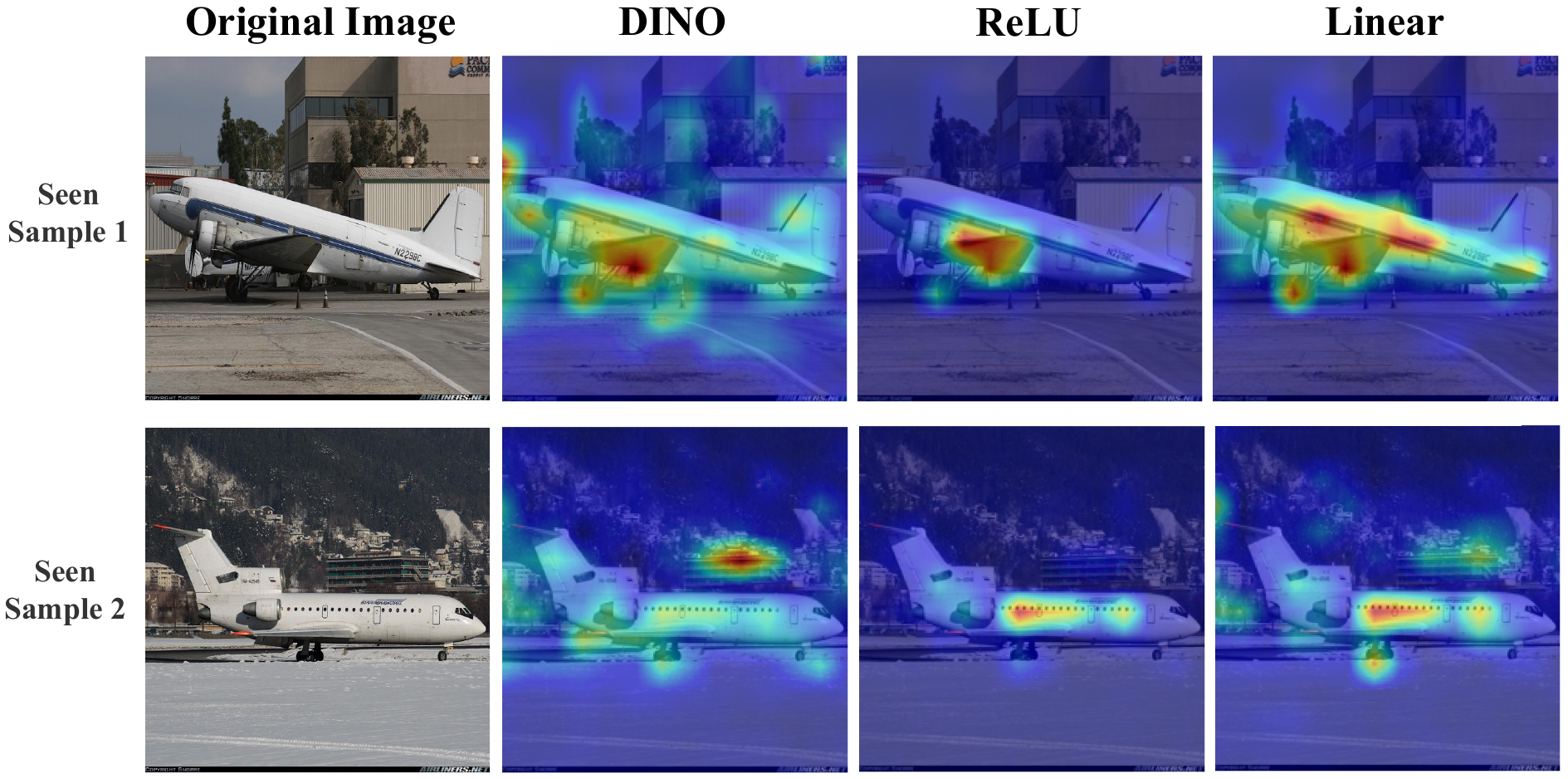}
\caption{\textbf{FGVC-Aircraft Seen Categories.}}
\label{fig:comparison2}
\end{minipage}
\end{figure*}

\begin{figure*}[t]
\centering
\begin{minipage}[b]{0.49\linewidth}
\centering
\includegraphics[width=\linewidth]{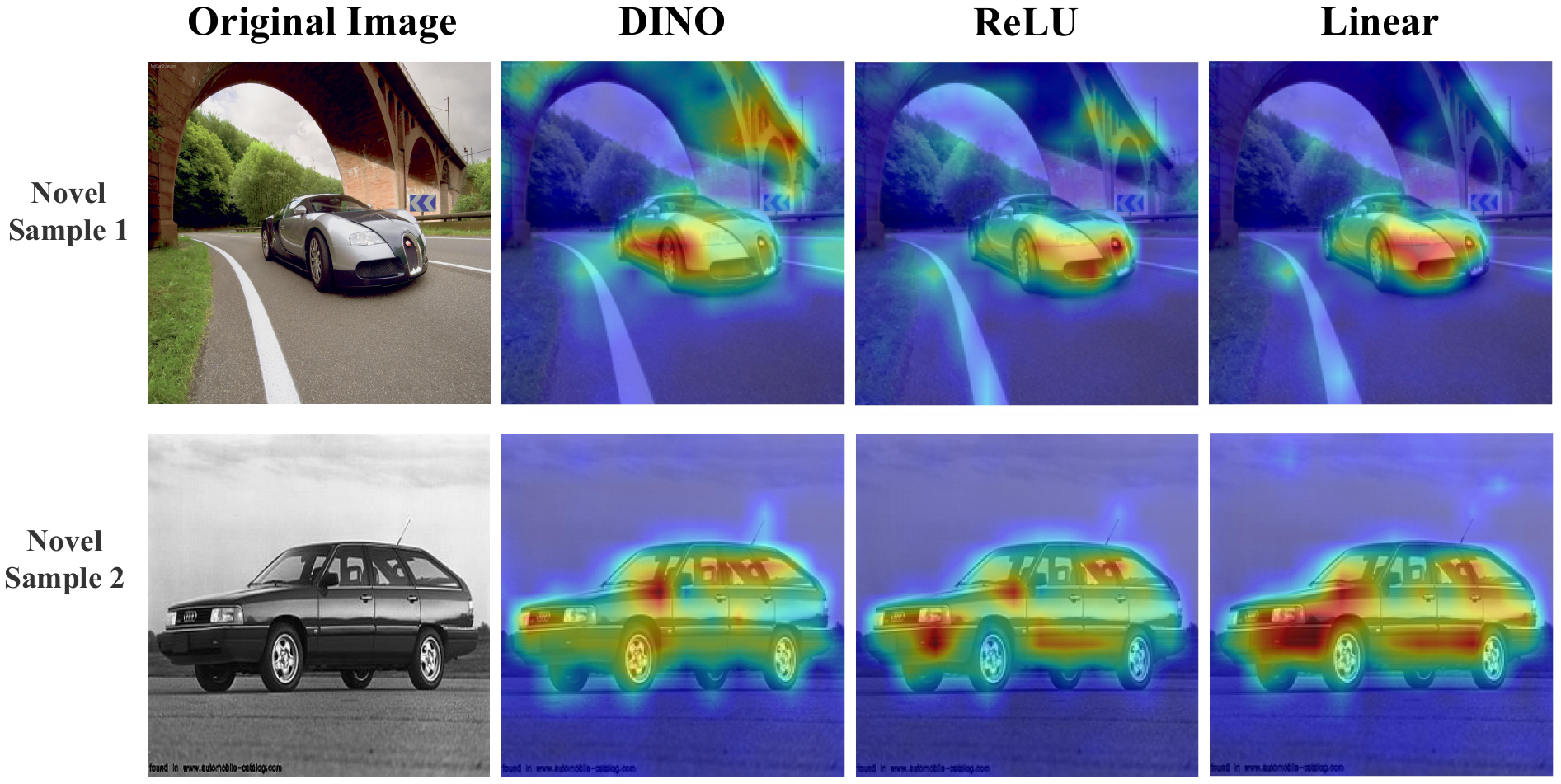}
\caption{\textbf{Stanford Cars Novel Categories.}}
\label{fig:comparison3}
\end{minipage}
\hfill
\begin{minipage}[b]{0.49\linewidth}
\centering
\includegraphics[width=\linewidth]{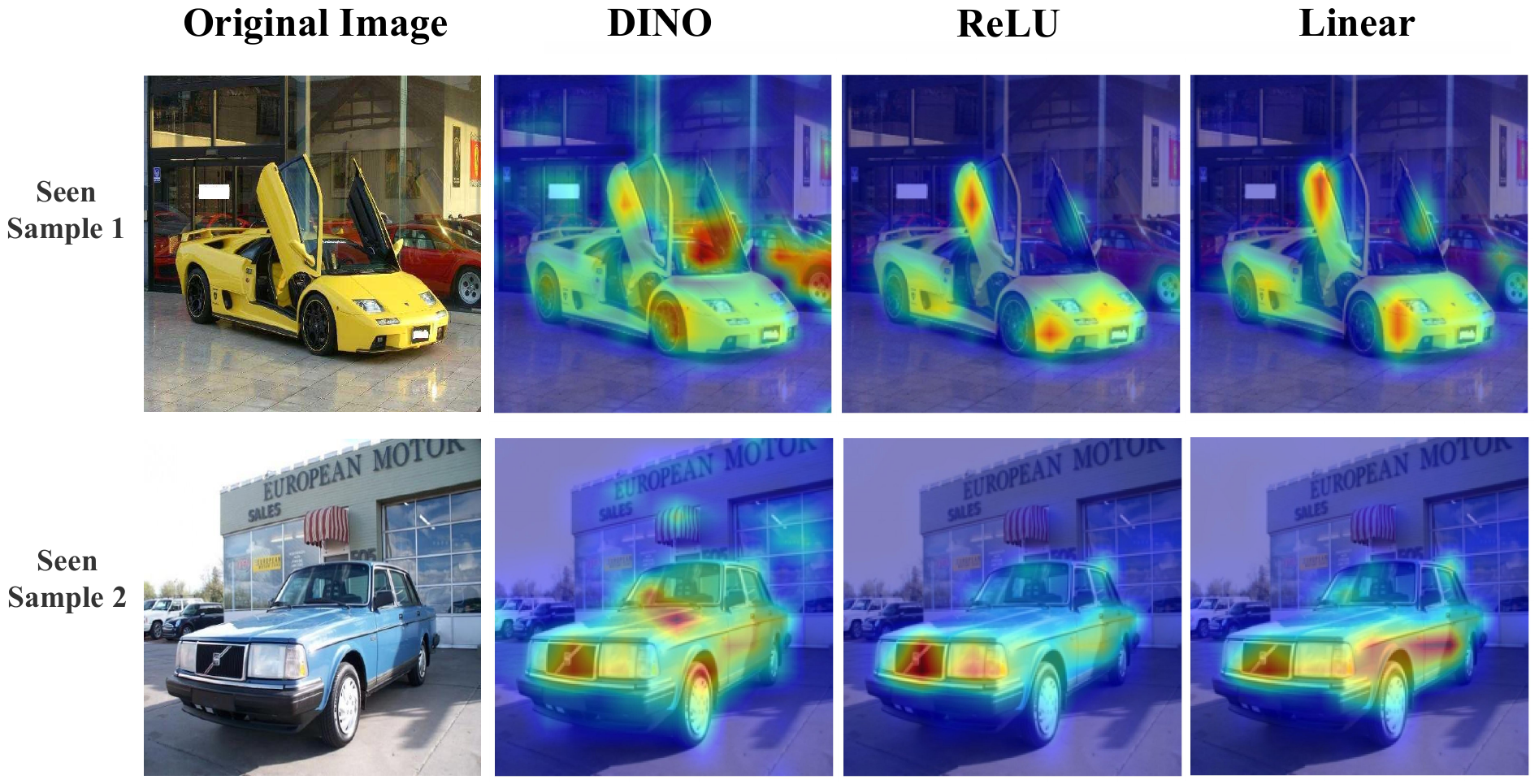}
\caption{\textbf{Stanford Cars Seen Categories.}}
\label{fig:comparison4}
\end{minipage}
\end{figure*}

\subsubsection{The scaling factor of distribution alignment}
The scaling factor $\mathit{s}_d$ influences the model’s focus on global class-level information during training as well as its reliance on balanced learning. In practice, it is difficult to accurately capture the true data distribution; thus, a uniform distribution is often employed to balance learning across classes, ensuring that underrepresented categories can still be improved. Previous methods have addressed this issue through a maximum-entropy regularization term, which encourages balanced learning, but its effectiveness remains limited. To mitigate the sensitivity of parameter tuning, we introduce distribution alignment as a complementary mechanism. As shown in Figure~\ref{fig:sens_align_scale}, performance remains relatively stable across scaling factors from 0.1 to 0.4. A slight decline is observed at $\mathit{s}_d = 0.4$, particularly in novel-category accuracy. These results suggest that although the model is not highly sensitive to $\mathit{s}_d$, excessively large $\mathit{s}_d$ can hinder generalization.

\subsection{Visualization}

We also provide attention maps to show why LinearAdapter outperforms ReluAdapter by visualizing the [CLS] token's self-attention patterns (averaged all attention heads in the final block). The visualization leverages 8 challenging images from Stanford Cars and Aircraft datasets that include architectural distractions, color variations, and multiple similar objects, providing the robust evaluation across both seen and novel classes. As shown in Figure~\ref{fig:comparison1}-\ref{fig:comparison4}, LinearAdapter demonstrates a superior capability in capturing more complete patterns of objects compared to ReluAdapter, which is highly beneficial for the GCD task.

\subsection{Incorporating Alternative Activations}
In the previous section, we examined ReLU and its variants, including Leaky ReLU, Threshold ReLU, and ELU, to assess whether the sparsity induced by ReLU constrains the model's adaptability. In this section, we extend the analysis to other commonly used activation functions and investigate how replacing ReLU with these alternatives influences model performance. We begin by providing detailed descriptions and comparisons of the activation functions under consideration:

\subsubsection{Sigmoid}
The Sigmoid activation function~\cite{rosenblatt1958perceptron}, commonly used in binary classification and probabilistic tasks, maps inputs to a range between $0$ and $1$, making it well-suited for modeling probabilities. It is defined as: 
\begin{equation}
\sigma(x) = \frac{1}{1 + e^{-x}},
\end{equation}
While Sigmoid is useful in output layers, it has two key limitations. First, its non-zero-centered outputs introduce gradient bias during backpropagation, which can slow convergence. Second, the gradient saturation occurs for large inputs, where $\sigma(x)$ approaches $0$ or $1$.

\begin{table*}[t]
\centering
\setlength{\tabcolsep}{2.0pt}
\caption{Detailed property comparisons of different activation functions}
\scalebox{1.09}{
\begin{tabular}{lccccc}
\hline
Activation & Output Range & Zero-Centered & Gradient Saturation & Computational Cost & Derivative Properties \\
\hline
Linear & (-$\infty$,$\infty$) & Yes & No & Very Low & Constant gradient \\
Sigmoid & (0, 1) & No & Yes & Medium & Smooth, max$=0.25$ at 0 \\
Tanh & (-1, 1) & Yes & Yes & Medium & Smooth, symmetric \\
ReLU & [0, $\infty$) & No & Only for $x < 0$ & Low & Discontinuous at 0 \\
Swish & (-0.3, $\infty$) & No & Partial & Medium & Smooth, non-monotonic \\
GeLU & (-0.28, $\infty$) & No & Partial & High & Smooth, continuous \\

\hline
\end{tabular}}
\label{table:activation_property}
\end{table*}

\begin{table*}[t]
\begin{center}
\renewcommand{\arraystretch}{0.75}
\setlength{\tabcolsep}{2.5pt}
\caption{{Evaluation results on three fine-grained datasets}. All experimental results were obtained from three runs of the experiment, with the corresponding mean and variance calculated. Bold values represent the best results.}
\label{table:activation}
\scalebox{1.05}{
\begin{tabular}{lcccccccccc}
\hline\noalign{\smallskip}
\multirow{2}{*}{\textbf{Method}}& \multicolumn{3}{c}{CUB200} & \multicolumn{3}{c}{FGVC-Aircraft} & \multicolumn{3}{c}{Stanford Cars} & \multirow{2}{*}{\textbf{Avg.}}\\
\cmidrule(lr){2-4}\cmidrule(lr){5-7}\cmidrule(lr){8-10}
& \cellcolor{blue!10} All & Seen & Novel & \cellcolor{blue!10} All & Seen & Novel & \cellcolor{blue!10} All & Seen & Novel \\
\noalign{\smallskip}
\hline
\noalign{\smallskip}
Linear & \cellcolor{blue!10} \textbf{68.77{\scriptsize $\pm$1.74}} & \underline{74.81{\scriptsize $\pm$0.48}} & \textbf{65.76{\scriptsize $\pm$2.82}} & \cellcolor{blue!10} \textbf{58.88{\scriptsize $\pm$0.76}} & \textbf{65.43{\scriptsize $\pm$0.44}} & \textbf{55.61{\scriptsize $\pm$0.94}} & \cellcolor{blue!10} \textbf{62.10{\scriptsize $\pm$0.83}} & \underline{78.24{\scriptsize $\pm$0.19}} & \textbf{54.31{\scriptsize $\pm$1.31}} & \cellcolor{blue!10}\textbf{64.88}\\
\noalign{\smallskip}
Swish & \cellcolor{blue!10} \underline{67.60{\scriptsize $\pm$0.88}} & \textbf{75.34{\scriptsize $\pm$0.90}} & 63.73{\scriptsize $\pm$1.37} & \cellcolor{blue!10} 56.08{\scriptsize $\pm$1.02} & 64.39{\scriptsize $\pm$0.85} & 51.93{\scriptsize $\pm$1.13} & \cellcolor{blue!10} \underline{59.81{\scriptsize $\pm$0.98}} & \textbf{78.83{\scriptsize $\pm$1.49}} & \underline{50.62{\scriptsize $\pm$2.01}} & \cellcolor{blue!10}\underline{63.15}\\
\noalign{\smallskip}
ReLU & \cellcolor{blue!10} 67.48{\scriptsize $\pm$0.74} & 73.91{\scriptsize $\pm$1.22} & \underline{64.26{\scriptsize $\pm$1.71}} & \cellcolor{blue!10} \underline{56.85{\scriptsize $\pm$0.74}} & \underline{65.39{\scriptsize $\pm$1.65}} & \underline{52.58{\scriptsize $\pm$0.57}} & \cellcolor{blue!10} 58.73{\scriptsize $\pm$1.53} & 76.66{\scriptsize $\pm$0.97} & 50.07{\scriptsize $\pm$2.45} & \cellcolor{blue!10}62.88\\
\noalign{\smallskip}
GeLU & \cellcolor{blue!10} 67.01{\scriptsize $\pm$0.76} & 74.40{\scriptsize $\pm$0.31} & 63.32{\scriptsize $\pm$1.03} & \cellcolor{blue!10} 56.37{\scriptsize $\pm$1.99} & 65.01{\scriptsize $\pm$0.42} & 52.05{\scriptsize $\pm$2.81} & \cellcolor{blue!10} 59.00{\scriptsize $\pm$1.22} & 76.58{\scriptsize $\pm$2.42} & 50.52{\scriptsize $\pm$2.30} & \cellcolor{blue!10}62.70\\
\noalign{\smallskip}
Tanh & \cellcolor{blue!10} 66.42{\scriptsize $\pm$1.32} & 72.98{\scriptsize $\pm$1.63} & 63.14{\scriptsize $\pm$1.15} & \cellcolor{blue!10} 55.91{\scriptsize $\pm$1.62} & 63.81{\scriptsize $\pm$0.26} & 51.96{\scriptsize $\pm$2.53} & \cellcolor{blue!10} 58.66{\scriptsize $\pm$0.81} & 75.69{\scriptsize $\pm$1.48} & 50.43{\scriptsize $\pm$1.01} & \cellcolor{blue!10}62.11\\
\noalign{\smallskip}
Sigmoid & \cellcolor{blue!10} 65.38{\scriptsize $\pm$1.01} & 72.36{\scriptsize $\pm$1.18} & 61.88{\scriptsize $\pm$0.92} & \cellcolor{blue!10} 51.46{\scriptsize $\pm$0.55} & 61.88{\scriptsize $\pm$0.42} & 46.26{\scriptsize $\pm$0.63} & \cellcolor{blue!10} 55.97{\scriptsize $\pm$0.69} & 71.48{\scriptsize $\pm$1.60} & 48.48{\scriptsize $\pm$0.26} & \cellcolor{blue!10}59.46\\
\noalign{\smallskip}
\hline
\end{tabular}
}
\end{center}
\end{table*}

\subsubsection{Tanh}
The Tanh (hyperbolic tangent) activation function~\cite{lecun2002efficient} is commonly used due to its zero-centered output, which can improve training dynamics. It is defined as:
\begin{equation}
\tanh(x) = \frac{e^x - e^{-x}}{e^x + e^{-x}} = \frac{\sinh(x)}{\cosh(x)},
\end{equation}
where $\sinh(x)$ and $\cosh(x)$ are the hyperbolic sine and cosine functions, respectively. Tanh maps inputs to the range $(-1, 1)$, providing outputs centered around zero, which helps balance gradients during training and can accelerate convergence. However, Tanh suffers from the gradient saturation: for inputs with large magnitudes, $\tanh(x)$ approaches $\pm 1$, causing its derivative to shrink towards zero, which could hinder the learning capacity.

\subsubsection{Swish}
The Swish activation function~\cite{ramachandran2017searching} is a smooth, non-linear function. It is defined as:
\begin{equation}
\text{Swish}(x) = x \cdot \sigma(x) = \frac{x}{1 + e^{-x}},
\end{equation}
where $\sigma(x)$ is the sigmoid function. 
Swish introduces a continuous and differentiable gating mechanism, smoothly suppressing negative inputs while preserving positive ones. Unlike ReLU, which abruptly discards negative values, Swish allows small negative activations, enabling richer feature representation.

\subsubsection{GeLU}
The GeLU (Gaussian Error Linear Unit) activation function~\cite{hendrycks2016gaussian} also offers a smoother and more flexible alternative to ReLU. Formally, GeLU is defined as:
\begin{equation}
\text{GeLU}(x) = x \cdot \Phi(x) = x \cdot \frac{1}{2} \left[1 + \text{erf}\left(\frac{x}{\sqrt{2}}\right)\right],
\end{equation}
where $\Phi(x)$ is the cumulative distribution function (CDF) of the standard normal distribution, and $\text{erf}(\cdot)$ is the error function. GeLU combines the strengths of both ReLU and sigmoid-like functions. Its smooth, non-linear form improves gradient flow, while its non-monotonic behavior allows for more complex transformations.

\subsubsection{Performance Comparison}

From the comparisons of these properties in Table~\ref{table:activation_property}, it can be observed that most activation functions introduce certain feature transformations that enhance the non-linear expressiveness. Nevertheless, such benefits typically come at the cost of compromising the output range, leading to value compression and gradient saturation. In the context of adapters, however, the advantage of introducing such non-linearity and sparsity remains unclear, since the setting does not involve training a randomly initialized model end-to-end, but rather adapting to small-scale downstream datasets. When replacing ReLU with alternative activation functions such as Sigmoid, Tanh, Swish, and GeLU, we observe no clear performance gains and in some cases even a decline compared to ReLU across all three datasets, as shown in Figure~\ref{table:activation}. Specifically, sigmoid shows a considerable performance gap compared with other activation functions. On FGVC-Aircraft and Stanford Cars, Linear exceeds the others by over 2\%, suggesting that activation functions in these cases hinder rather than enhance performance.

Upon closer comparison, it becomes evident that the model’s performance is more closely associated with the output range and gradient saturation of the activation functions, while the specific form of non-linear transformation appears to play a relatively minor role such as Swish and GeLU. For instance, Sigmoid performs noticeably worse than the other activation functions due to its severe gradient saturation and strong compression of features. In contrast, as the output range and gradient saturation improve with Tanh, Swish, and GeLU, the performance exhibits a clear enhancement. Interestingly, simply removing the activation functions, i.e. adopting the linear case, preserves the features, provides the widest output range and stable gradients, and achieves better adaptation than adapters with non-linear activations.

\section{Conclusion and Limitation}
This paper introduces LAGCD, a simple yet effective adapter-based tuning method for Generalized Category Discovery (GCD). LAGCD enhances the model’s adaptability by integrating a lightweight linear adapter into each ViT block, achieving superior performance over prior methods relying on partial fine-tuning or visual prompt tuning. Our systematic analysis reveals that ReLU-induced feature sparsity and biased seen/novel predictions hinder overall generalization in GCD. To address these issues, we eliminate the activation function in adapters and optimize an auxiliary distribution alignment loss, which together facilitate more effective adaptation and balanced category learning. Extensive experiments demonstrate that our method achieves state-of-the-art performance across diverse fine-grained and generic datasets. This work highlights the overlooked role of activation functions in adapter design and introduces a method that establishes a strong baseline.

Although our method achieves competitive performance with significantly fewer parameters, along with improved stability and adaptability, some limitations remain to be addressed. For instance, while the removal of ReLU helps alleviate feature sparsity and enhances generalization in GCD, the general applicability of this design across diverse tasks has yet to be thoroughly investigated.

\ifCLASSOPTIONcaptionsoff
  \newpage
\fi

\normalem
\bibliographystyle{IEEEtran}

\bibliography{reference}

\end{document}

%% file: math_commands.tex
\usepackage{amsmath,amsfonts,bm}

\def\eqref#1{equation~\ref{#1}}

\def\1{\bm{1}}

\DeclareMathAlphabet{\mathsfit}{\encodingdefault}{\sfdefault}{m}{sl}
\SetMathAlphabet{\mathsfit}{bold}{\encodingdefault}{\sfdefault}{bx}{n}

